\documentclass[journal]{IEEEtran}
\IEEEoverridecommandlockouts
\usepackage{cite}
\usepackage{amsmath,amssymb,amsfonts}
\usepackage{amsmath}
\usepackage{algorithmic}
\usepackage{graphicx}
\usepackage{textcomp}
\usepackage{xcolor}
\usepackage{indentfirst}
\usepackage{amsmath}
\usepackage{float}
\usepackage{cite} 
\usepackage{booktabs}
\usepackage{comment}
\usepackage{array}
\title{\LARGE \bf Encoding Biomechanical Energy Margin into Passivity-based Synchronization for Networked Telerobotic Systems}

\author{Xingyuan Zhou$^*$, \textit{IEEE Student Member}, Peter Paik, \textit{IEEE Student Member}, \\ S. Farokh Atashzar{}, \textit{IEEE Senior Member}

\thanks{This work was supported by the US National Science Foundation under grants no \#2121391 and \#2208189. The work is also partially supported by NYUAD Center for Artificial Intelligence and Robotics (CAIR) award \# CG010.  The authors would like to acknowledge the support from
MathWorks.\textit{({$^*$}Corresponding Author: Xingyuan Zhou.)} \par
Xingyuan Zhou and Peter Paik are with the Department of Electrical and Computer Engineering, New York University (NYU), New York, NY
11201, USA (email: xz3428@nyu.edu; hsp287@nyu.edu). \par
At the time of conducting the research, Farokh Atashzar was affiliated with NYU.}
    }

\begin{document}
	
	\maketitle
	\thispagestyle{plain}
	\pagestyle{plain}
	\bstctlcite{IEEEexample:BSTcontrol}

\begin{abstract}
{Maintaining system stability and accurate position tracking is imperative in networked robotic systems, particularly for haptics-enabled human-robot interaction. Recent literature have integrated human biomechanics into the stabilizers implemented for teleoperation, enhancing force preservation while guaranteeing convergence and safety. However, position desynchronization due to imperfect communication and non-passive behaviors remains a challenge. This paper proposes a Two-Port Biomechanics-aware Passivity-based Synchronizer and Stabilizer (TBPS\textsuperscript{2}). This stabilizer optimizes position synchronization by leveraging human biomechanics while reducing the stabilizer’s conservatism in its activation. We provide the mathematical design synthesis of the stabilizer and the proof of stability. We also conducted a series of grid simulations and systematic experiments and compared the performance with state-of-the-art solutions regarding varying time delays and environmental conditions. Results show that TBPS\textsuperscript{2} significantly reduces position drift and velocity deflection, demonstrating superior performance over existing state-of-the-art solutions. The proposed stabilizer is effective for various telerobotic applications requiring precise position synchronization.}

\end{abstract}


\section{Introduction}

{In the field of physical human-robot interaction, especially for teleoperation scenarios, two criteria should be prioritized: ensuring the safety of the interaction and optimizing energy transfer from human biomechanics to robotic mechanics. These criteria deeply relate to stability and transparency--two essential challenges in haptics and teleoperation. For instance, low-quality network\cite{Farajiparvar2020} services such as time delays, packet losses, and jitter between the two ends of the haptic interfaces (e.g., the leader and the follower robot) can disrupt the system's stability and transparency due to the non-passive networked coupling. Furthermore, internal electromechanical failures within the teleoperation platform, such as encoder faults, noise, and measurement imperfections, also raise additional challenges to maintain safety, transparency, and overall system reliability. }

{Therefore, vast research has focused on designing stabilization solutions for telerobotic systems to ensure stability, specifically for addressing non-passivity caused by communication delay. Some popular stabilizing algorithms examples are: wave variable control \cite{sun2015wave}, time-domain passivity control \cite{Zhou2024,ryu2010passive,shahbazi2018position,RyuPro2007}, scattering control \cite{anderson1988bilateral}, small-gain control \cite{atashzar2016small}, and passive set-position modulation \cite{LeeTRO}. In most teleoperation tasks, the remote environment is assumed to be passive environment, meaning all objects the robot interacts with are static and do not destabilize the system. Therefore, the passive-environment assumption underpins the design of nearly all stabilizers for telerobotic systems.}

{ However, this assumption does not hold for more complex applications such as telerobotic rehabilitation and telesurgery. In telerobotic rehabilitation, a remote therapist provides assistive forces to augment the patient's impaired motor functions, which is inherently non-passive behavior according to passivity control theory. This non-passivity behavior from the environment will challenge the performance of traditional stabilizers, especially those designed under the assumption of environmental passivity, as discussed in \cite{sfaTRO}. Similarly, telerobotic surgery, particularly on operating on organs like the beating heart, also breaks this passive environment assumption. More details can be found in \cite{sarmad2021} and references therein.}

{Recently, researchers tried to tackle the non-passive environment issue through different approaches, for example, by proposing stabilization methods that do not rely on passivity behaviors, like nonlinear IOS small-gain stabilization\cite{atashzar2016small,Polushinsmall,Polushinsmall2}. Alternatively, some researchers propose shifting the entire stabilization effort to the leader's side control, which seeks to minimize the non-passive effect from the by solely adjusting the force reflection based on the dynamic responses of the leader's operator \cite{atashzar2016passivity}.
Although the mentioned solutions have helped to relax the environment passive assumption, they also introduce drawbacks. Both approaches might degrade the quality of human-telerobot interaction by significantly altering how information transfers between the user and the robot, such as feedback force transparency and position tracking.}

{Our previous research highlighted the potential benefit of incorporating human biomechanics into stabilizer design to enhance feedback force transparency. By harnessing the energetic ability of human biomechanics to absorb interactional energy, which is a concept related to ``Excess of Passivity,"  we can transfer some stabilization burdens from the stabilizer to the human user \cite{zhouToH2023}\cite{Zhou2023UpperlimbGM}. This shift utilizes the inherent energy margin of human biomechanics, leading to reduced reliance on the stabilizer and improved fidelity of feedback forces.}

{However, position synchronization remains a challenge in teleoperation scenarios. 
In practice, the human-telerobot position synchronization has been a topic of concern due to the following reasons: (a) several categories of existing stabilizers such as wave variable control \cite{sun2015wave}, scattering control \cite{anderson1988bilateral}, and initial designs of TDPA \cite{ryu2010passive} distribute the stabilization load between velocity tracking and force tracking characteristics of the system. The velocity tracking error over time would result in an accumulative position drift, which can eventually fail to satisfy tasks that require accurate position tracking performance. This concept has been investigated in the literature, and thus, variations of solutions have been proposed to address the position synchronization issue \cite{artigas2010position,OMalleyTDPA,tdpa-cf,tdpa-ps,tdpa-eng}.}

{For example, in \cite{artigas2010position}, the authors exploit the discoverable positive energy gaps in real-time and utilize them in the format of energy pockets (representing the margins of stability) to reduce position drift without violating the passivity-based stability condition. An improvement to this method with force and torque smoothening is given in \cite{tdpa-ps}. In addition, other state-of-the-art solutions that reduce the position drift through chattering reduction \cite{tdpa-cf}, and through energy injection \cite{tdpa-eng} have been recently introduced. Even though these methods can reduce the position drift issue, they each have their own drawbacks that will be examined later in the results.}


{One of the reasons the preivous proposed stabilizers are not ideal is that such algorithms ignore the inherent energy absorption of the human operator's biomechanics in the design of the stabilization method. Human operator's biomechanics can be viewed as the biological energy margin in the control loop, which can be taken into account in the design of stabilizers to balance the excessive energy and reduce the need for injecting damping into the control loop of human-robot interaction. This topic has been discussed in our previous research, which can be found in \cite{atashzar2016passivity,zhouacc2024,paiktim2025}. The corresponding family of stabilizers can guarantee stability while maintaining high transparency by avoiding over-activation. 
}

{In this paper, for the first time, we propose a novel biomechanics-aware stabilization module that can take into account the biomechanical energy margin of human biomechanics to (a) recover position synchronization, (b) guarantee “minimal L2 stability” while reducing the conservatism,  and (c) significantly reduce the activation of the stabilizer, all concurrently. The stabilizer is named Two-Port Biomechanics-aware Passivity-based Synchronizer and Stabilizer (TBPS\textsuperscript{2}).

In order to validate the performance of the proposed TBPS\textsuperscript{2}, we conducted a systematic simulation and comprehensive experiments. The results show that the performance of the TBPS\textsuperscript{2} is superior to the other state-of-the-art stabilizers considering position drift compensation as well as the velocity deflection and activation of the stabilizer. This significant improvement can benefit various practical applications such as telerobotic surgery or telerobotic rehabilitation.\par
The rest of this paper is organized as follows:  in Section II, the preliminaries are provided, the controller design of the TBPS\textsuperscript{2} is presented in Section III, and the simulation verification results and the systematic experiments are given in Section IV and V, respectively. Finally, the paper is concluded in Section VI.

\begin{figure}[htbp]
\centerline{\includegraphics[width=0.46\textwidth]{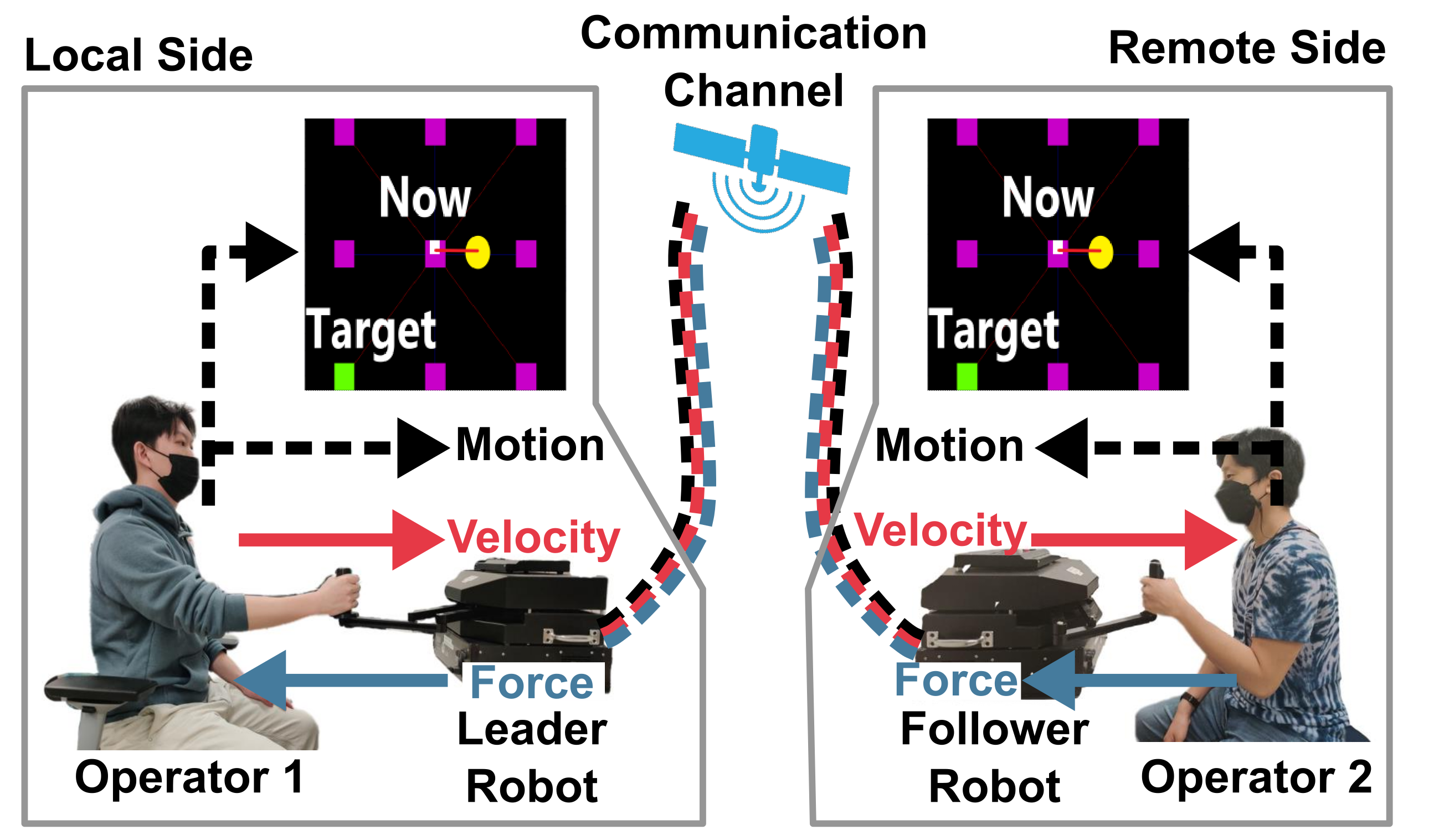}}
\caption{Block diagram of a Leader-Follower telerobotic architecture with telerehabilitation scenario. The virtual environment is shared between the leader and follower side, where the white square and yellow circle represent leader operator and follower operator movements, respectively.}
\label{fig1}
\end{figure}

\section{Background}

\subsection{System Description}

{As shown in Fig. \ref{fig1}, a two-channel architecture is considered for the networked telerobotic system to exchange the velocity and force information bilaterally. Thus, the leader robot measures the operator's motion and communicates to the follower side through a delayed communication network to conduct a task in a remote environment. The follower robot measures the interactive environmental forces and delivers them to the leader robot to render a haptic perception during task execution.}

{In this paper, all analyses are conducted after applying an optimized Lawrence two-channel bilateral Leader-Follower telerobotic architecture \cite{sfaTRO,ryu2004stable}. Accordingly, the residual interconnected dynamics are shown in Fig. \ref{fig2}.}
\begin{figure}[htbp]
\centerline{\includegraphics[width=0.48\textwidth]{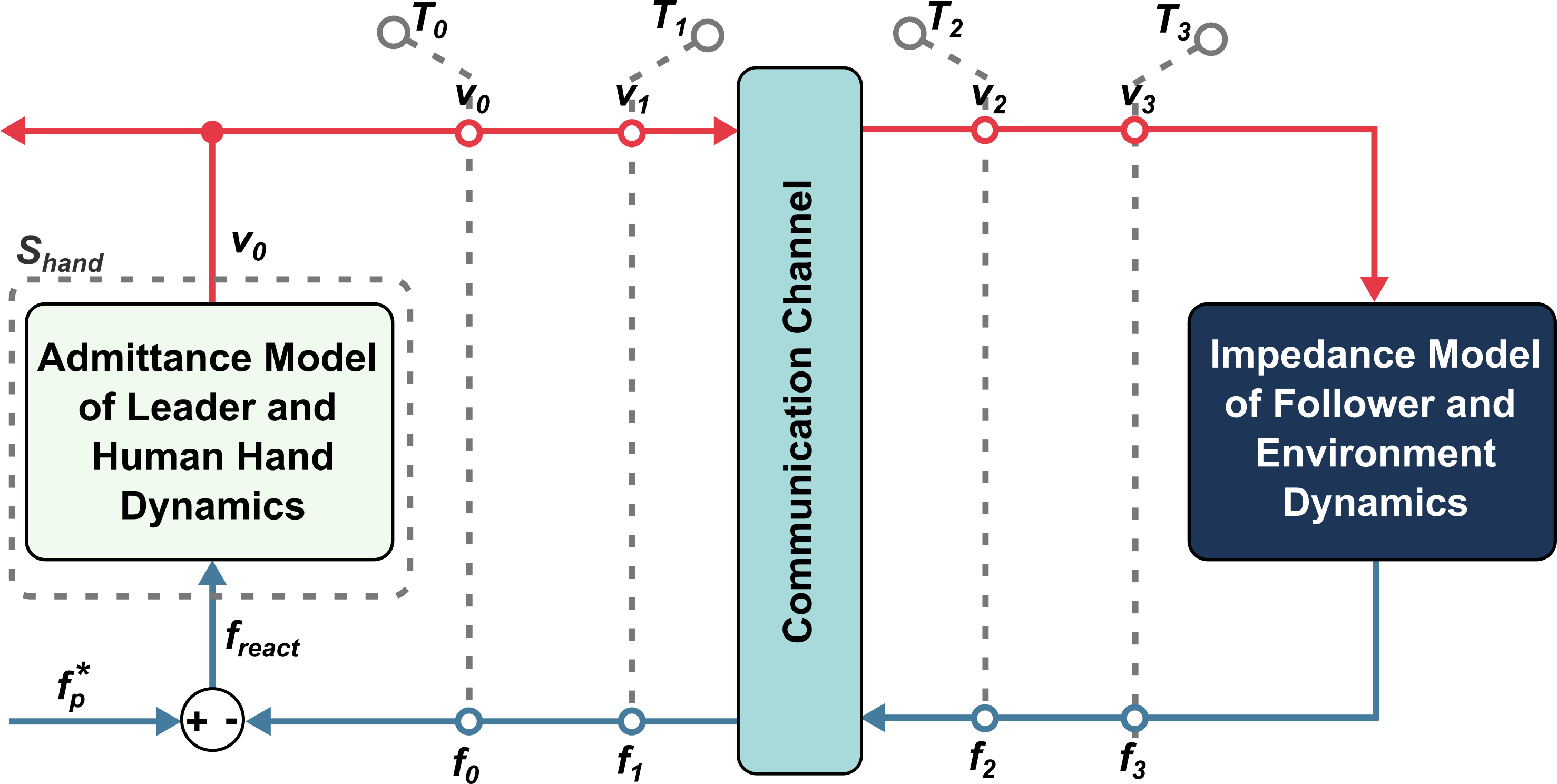}}
\caption{Overall schematic of a two-channel telerobotic interconnection.}
\label{fig2}
\end{figure}

{In an ideal scenario without communication delays and non-passive sources, this architecture would result in a negative interconnection of two passive subsystems. According to Weak Passivity Control Theory, such a configuration would inherently maintain passivity, thereby ensuring stability \cite{vidyasagar2002nonlinear}. At the same time, the architecture would guarantee transparency, which means that it would hypothetically guarantee perfect force and velocity tracking \cite{atashzar2016passivity}. Such that $f_{3}$ = $f_{0}$ and $v_{0}$= $v_{3}$. This means that the interaction force $f_3$ exerted by the follower robot would perfectly match the force $f_0$ rendered to the operator by the leader robot. Similarly, the velocity of the leader robot $v_0$ would be perfectly matched by the velocity of the follower robot as $v_3$.}

{Moreoever, as depicted in Fig. \ref{fig2}, the left side of the communication channel represents the admittance model of the operator’s reaction dynamics. Here, $v_1$ denotes the velocity transmitted to the follower side, while $f_1$ represents the force received from the follower side via the communication channel. The impedance model of the environmental behavior is illustrated on the right side of the communication channel. Within this model, $v_2$ at the follower side is the velocity received from the leader robot through the forward communication channel, and $f_2$ is the force transmitted from the follower side to the leader robot via the backward communication channel.}

{Regarding the behavior of human biomechanics in the loop of telerobotic system, the operator's side interactional force $f_0$ can be decoupled into two parts. First, a component that generates the active motion $f^*_p$, often named ``the exogenous force'' applied by muscle contractions. Second is an inherent reactive component of human biomechanics during physical human-robot interaction $f_{react}$, which represents the instantaneous biomechanical resistive forces in response to the robot motion, as can be seen in (\ref{eq:1}).}

\begin{equation}
f_{0} = f_{p}^{*} - f_{react} \quad\text{and}\quad  f_{react} = z_{p}(v_{0},t).
\label{eq:1}
\end{equation}

{The reactive biomechanical force is modeled using $z_{p}(v_{0},t)$, which is a non-autonomous nonlinear impeding dynamical behavior of the operator. Considering Fig. \ref{fig2}, we introduce the terminology of “Observation Terminals” to organize the mathematical derivations and explanations. In this regard, $v_i$ and $f_i$ are the velocity and force at the $i^{th}$ observation terminal denoted as $T_i$ in Fig. \ref{fig2}. As a result, the energy measured at Terminal $T_i$ in the discrete-time domain is:
}
\begin{equation}
       E_{T_i}(k) = \Delta T\sum_{k} f_{i}(k)v_i(k),   \\
\label{eq:2}
\end{equation}
where the sampling rate is defined as $\Delta T$ and the discrete time step is `$k$' where $k\Delta T=t$.

\subsection{Conventional Time Domain Passivity Control}
{Time Domain Passivity Approach (TDPA) \cite{RyuPro2007} is one of the commonly used stabilizers in the literature, which defines a family of controllers that (a) observe the flow of interactional energy when monitoring the validity of a passivity condition and (b) adaptively inject damping to dissipate the observed excessive energy and thus guarantee passivity and eventually stability. Therefore, there are two necessary components in the family of TDPA stabilizer architecture:}

\begin{itemize}
\item First, is the passivity observer (PO), which keeps track of the interactional energy flow in real-time at each side of the communication channel. It is formulated based on the observed energy information to validate the passivity condition of the networked telerobotic system.
\item Second, is the Passivity Controller (PC) that can be placed on either or both sides of the communication channel. PCs are adaptive virtual dampers that modify the interactional velocities and forces to dissipate the extra energy observed by POs and thus guarantee the system passivity and stability.\par
\end{itemize}
In order to have a better understanding of the functionality of the family of TDPA stabilizers, the mathematical definitions of the passive networked system are given below based on \cite{Hill1977,Jazayeri2013,vidyasagar2002nonlinear,2022Ryu}. \par

\textbf{\textit{Definition 1 - One Port System Passivity:}}
A one-port network system $\Phi _1(k)$, with input of $U(k)$, and output of $Y(k)$, and initial energy $E(0)$, is passive if and only if $E_{\Phi _1}(k) + E(0) \geq 0$ for all $k > 0 $, where $E_{\Phi _1}(k)$ is defined as the energy of the system and can be calculated as:
\begin{equation}\label{4}
E_{\Phi _1}(k) = \Delta T\sum_{k} U(k)^TY(k).
\end{equation}

\textbf{\textit{Definition 2 - Multi-Port System Passivity:}}
Generalizing (\ref{4}), it can be mentioned that an M-port system $\Phi _M(k)$ with input vector $U(k)$, output vector $Y(k)$, and initial energy $E(0)$, is passive if and only if $E_{\Phi _M}(k) + E(0) \geq 0$ for all $k > 0$, where $E_{\Phi _M}(k)$ is defined as the energy of the M-port system and can be calculated as:

\begin{equation} \label{5}
E_{\Phi _M}(k) = \Delta T\sum_{i=1}^{M}\sum_{k} U_{i}(k)^T Y_i(k).
\end{equation}
In (\ref{5}), $<U_i,Y_i>$ is the input-output vector couple at port $i$. \\

\textbf{\textit{Definition 3 - Interconnected System Passivity: }}If there is an interconnected system comprising of \textit{j} arbitrarily connected subsystems $S$, the whole interconnection is passive if $E_{Net}(k) + E(0) \geq 0$ for all $k> 0$, where $E_{Net}(k) = E_{S_1}(k) + E_{S_2}(k) + E_{S_3}(k) ... E_{S_j}(k) $, and $E_{S_j}(k)$ is the $j^{th}$ subsystem's energy. In other word, the interconnection system is passive if the summation of the every subsystem's energy is greater than zero. In this context, $E_{Net}(k)$ is the total energy of the networked system. The reference of the interconnection system is illustrated at Fig.3.

\begin{figure}[htbp]
\centerline{\includegraphics[width=0.3\textwidth]{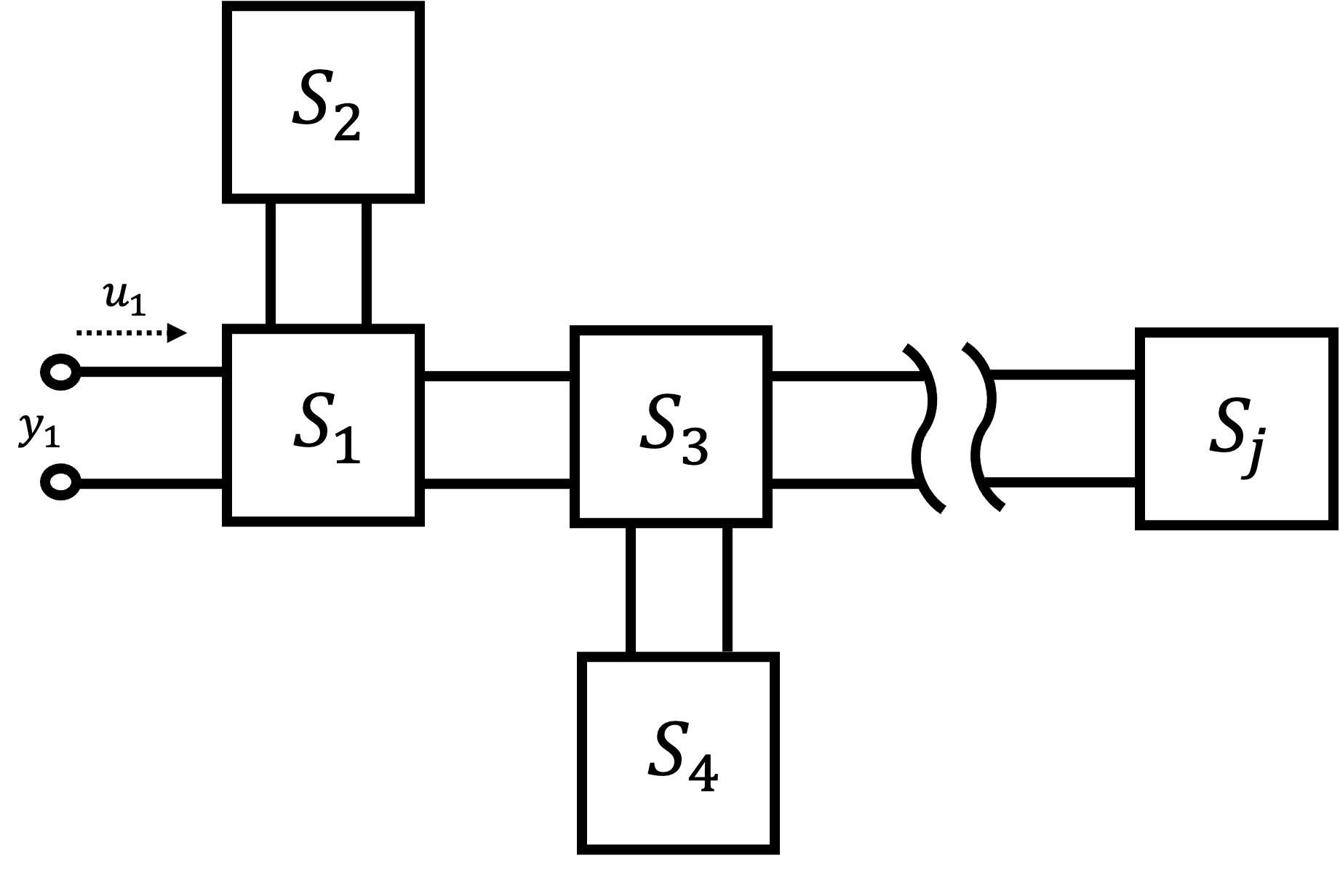}}
\caption{Interconnection system consisting of `j' subsystems}
\label{fig3}
\end{figure}

\textbf{\textit{Definition 4 - Output Strictly Passive System:}}
For any system with input $U(k)$ and output $Y(k)$, and initial energy $E(0)$, is considered as Output Strictly Passive system when 
\begin{equation}
\sum_{k}U(k)^TY(k)\Delta T + E(0) \geq  \xi\sum_{k}Y(k)^TY(k)\Delta T.
\end{equation}

If $\xi \geq 0$, then we call the system Output Strictly Passive (OSP) with EoP of $\xi$. If $\xi < 0$, then we call the system Output Non-Passive (ONP) with Shortage Of Passivity (SoP) of -$\xi$. If a system is strictly passive, then it is also asymptotically stable. An OSP system is also an L2 stable with finite L2 gain $\leq 1/\xi$ where $\xi$ is the EoP of the OSP system \cite{atashzar2016small}.\par

{\textbf{\textit{Definition 5: Relation between Passivity and Stability:}}
if the system is an Output Strictly Passive system with EoP of $\xi$, then the system is L2 stable with finite L2 gain $\leq 1/\xi$. For a more general form, we proof the relationship between passivity and stability in the continuous time domain. For the OSP system, we have:}
{
\begin{equation}
\int_{0}^{T} U^{T}(t) Y(t) \, dt +E(0) \geq  + \xi \int_{0}^{T} Y^{T}(t) Y(t) \, dt
\end{equation}
}
{
The condition can be rewritten as:
}
{
\begin{equation}
\int_{0}^{T} \left( U^{T}(t) Y(t) - \xi Y^{T}(t) Y(t) \right) dt \geq 0
\end{equation}}
{
assuming that the initial energy is zero $E(0)=0$. It can be shown that, algebraically:}
{
\begin{equation}
U^{T} Y - \xi Y^{T} Y = -\frac{1}{2\xi} (U - \xi Y)^{T} (U - \xi Y) + \frac{1}{2\xi} U^{T} U - \frac{\xi}{2} Y^{T} Y
\end{equation}}
{
such that the condition can be rewritten as:}
{
\begin{equation} \label{rewritten}
    \int_{0}^{T} \left( -\frac{1}{2\xi} (U - \xi Y)^{T} (U - \xi Y) + \frac{1}{2\xi} U^{T} U - \frac{\xi}{2} Y^{T} Y \right) dt \geq 0
\end{equation}}

{
also, since $ -\frac{1}{2\xi} (U - \xi Y)^{T} (U - \xi Y)$ is negative semi-definite, we know that }

\begin{equation}
\begin{split}\label{generalfinal}
&\int_{0}^{T} \left( \frac{1}{2\xi} U^{T} U - \frac{\xi}{2} Y^{T} Y \right) dt \\
&\geq \int_{0}^{T} \left( -\frac{1}{2\xi} (U - \xi Y)^{T} (U - \xi Y) + \frac{1}{2\xi} U^{T} U - \frac{\xi}{2} Y^{T} Y \right) dt
\end{split}
\end{equation}

{
as a result, combining (\ref{rewritten}) and (\ref{generalfinal}), for the OSP system, we have: }
{
\begin{equation}
    \int_{0}^{T} \left( \frac{1}{2\xi} U^{T} U - \frac{\xi}{2} Y^{T} Y \right) dt \geq 0
\end{equation}
}
{
which we can rewrite as 
}
{
\begin{equation}
\int_{0}^{T} Y^{T}(t) Y(t) \, dt \leq \frac{1}{\xi^2} \int_{0}^{T} U^{T}(t) U(t) \, dt
\end{equation}}
{
Therefore, the OSP system is L2 stable with a finite L2 gain:}
{
\begin{equation}
\| Y_{T} \|_{L_2} \leq \frac{1}{\xi} \| U_{T} \|_{L_2}
\end{equation}}

In the design of TDPA stabilizers, the definitions mentioned above are often utilized to observe and assess the passivity of the Two-port networked systems comprising three subsystems i.e., (a) human operator, (b) network, and (c) environment. As a result, the energy of a bilateral telerobotic system is formulated as $E_{Hu}(k) + E_{Com}(k) + E_{En}(k)$. It should be noted that the human operator’s biomechanics is commonly regarded as a strictly passive system \cite{sfaRALhip,Hogan2016ankle,atashzar2017grasp} thus we have $E_{Hu}(k)\geq 0$. As a result, a conservative passivity criterion (used in the literature for TDPA stabilizers) for two-port networked telerobotic systems in the presence of an unknown environment is:

\begin{equation}\label{6}
 E_{Com}(k) + E_{En}(k) \geq  0  ,  \text{ $k > 0 $}.
\end{equation}
Considering Fig. \ref{fig2} and (\ref{6}), it can be mentioned that $E_{En}(k)$ is equal to  $E_{T_3}(k)$ which is the energy at terminal $T_3$. Also, taking into account the respective forces and velocities and the corresponding directions at $T_1$ and $T_2$  terminal in Fig. \ref{fig2}, it can be shown that $E_{T_1}(k)=E_{com}(k)+ E_{T_2}(k)$, which means that $E_{com}(k)= E_{T_1}(k)-E_{T_2}(k)$. Thus, the passivity criterion given in (\ref{6}) can be re-written as 

\begin{equation}\label{eq:ovenergy}
    E_{T_1}(k)- E_{T_2}(k)+E_{T_3}(k) \geq 0.
\end{equation}
It should be also noted that the energy at each terminal can be split into positive and negative contributions \cite{artigas2010position}. The sign of the power indicates the direction of the flow at that time stamp. As a result, the accumulation of positive power can be defined as the energy flowing into an observation terminal, like $T_i$, and the accumulation of negative power can be defined as the energy flowing out of the terminal. Thus we have

\begin{equation}\label{9}
\begin{cases}
        E_{T_i}^{in}(k) = \Delta T\sum_{k} (f_{i}(k)v_i(k))^{in}, & \text{ $f_{i}(k)v_{i}(k) \geq 0$},\\\\       
     E^{out}_{T_i}(k) = \Delta T\sum_{k} (f_{i}(k)v_i(k))^{out}, & \text{ $f_{i}(k)v_{i}(k) < 0$},
  \end{cases}
\end{equation}

and

\begin{equation}
E_{T_i}(k) = E^{in}_{T_i}(k) + E^{out}_{T_i}(k). \quad\quad 
\end{equation}
The superscript notation "in" represented the input/positive term and the superscript "out" represented the output/negative term. Considering the equations above, the passivity condition from \eqref{eq:ovenergy} can be rewritten as:
\begin{equation}\label{10}
E_{T_1}^{in}(k) + E_{T_1}^{out}(k)- E_{T_2}^{in}(k) - E_{T_2}^{out}(k) + E_{T_3}^{in}(k) + E_{T_3}^{out}(k) \geq 0. 
\end{equation}
Combining (\ref{9}) and (\ref{10}), it can be mentioned that a telerobotic system, shown in Fig. \ref{fig2} is passive and thus stable, if 

\begin{equation}
\begin{split}\label{generalfinal1}
&\Delta T_L\sum_{k}{[(f_1(k)v_1(k))^{in} + (f_1(k)v_1(k))^{out}]} \\
&- \Delta T_F\sum_{k}[(f_2(k)v_2(k))^{in}- (f_2(k)v_2(k))^{out} + (f_3(k)v_3(k))^{in} \\
&+ (f_3(k)v_3(k))^{out}] \geq 0.
\end{split}
\end{equation}

$\Delta T_L$ and $\Delta T_F$ are the sampling duration at Leader and Follower side respectively, which usually is different in practice.
\subsection{Estimation of EoP}

In order to fully utilize the benefits of the human operator’s biomechanics, it is essential to quantify and track how much energy it can absorb during human-robot interaction. Recently, our team has investigated the mathematical foundation and the experimental procedure to estimate the EoP \cite{B_Oliver2023}\cite{A_Oliver2023MyoPassivity,atashzar2016passivity,atashzar2017grasp}. EoP is a nonlinear human biomechanical property related to how much interactional energy can be absorbed during human-robot interaction. By utilizing this information, even with the most relaxed condition we can still estimate the lower bound of the energy absorption capability of the human. In order to define EoP, below, we first provide the definition of Output Strictly Passive Systems.

 In regards to the teleoperation architecture, the OSP condition is evaluated for the admittance model of the operator's hand dynamics represented in Fig. \ref{fig2} as `$S_{hand}$'.\par

Therefore, for `$S_{hand}$', the OSP condition is given by:

\begin{equation}\label{13}
\sum_{k}f_{react}(k )^{T}v_{0}(k)\Delta T \geq  \xi\sum_{k}v_{0}(k)^Tv_{0}(k)\Delta T,
\end{equation}
where the EoP is defined by $\xi$. Specifically, the energy absorption capability of the hand `$E_{hand}(k)$' is described by the left-hand side of the inequality in \eqref{13}. Likewise, a lower-bound estimate of the energy absorption capability `$\hat{E}_{hand}(k)$' is given by the right-hand side of the inequality in \eqref{13}. In this regard, the estimation of the absorption capability is encoded in the EoP `$\xi$'. This is necessary since a pure measurement of $f_{react}$ is not available in practice when the operator is moving the robot handle (i.e., $f_{p}^{*}\neq0$).

To overcome this, an offline identification protocol before the experiment to calculate the EoP has been developed and discussed in previous publications \cite{atashzar2017grasp,sfaRALhip}. During the identification phase, the operator is asked to hold the robotic handle in a relaxed condition (i.e., applying minimal grasping force to the handle) \cite{atashzar2016passivity}. {Then, the robot applies a perturbation. The perturbation is a high-frequency(3Hz), short-distance(3cm) repeating motion, and the perturbation is applied in eight different compass directions.} During this phase, the operator's hand does not apply any exogenous force `$f_{p}^{*}=0$'; therefore, the `$f_{react}$' can be directly measured. This allows us to measure the $\xi_{relax}$ using \eqref{13}:




\begin{equation}\label{14}
\xi_{relax} = \frac{\sum_{k}f_{react}(k)^{T}v_{0}(k)\Delta T}{ \sum_{k}v_{0}(k)^{T}v_{0}(k)\Delta T}.
\end{equation}
The estimated $\xi_{relax}$ is the EoP of the human hand in a relaxed condition in one of the eight compass directions. This identification protocol is repeated for the other compass directions and also for when the hand is tightly grasping the handle of the robot. Then, through linear extrapolation, the $\xi$ value of the hand can be interpreted in real-time during the experiment based on the direction of interaction and grasp pressure. As a result, we can obtain a lower-bound estimate of the energy absorption capability of the hand `$\hat{E}_{hand}(k)$' without measuring $f_{react}$. Details can be found in \cite{Zhou2023UpperlimbGM}\cite{atashzar2017grasp}\cite{zhouiros2023}.


During the last five years, we have utilized the estimation of EoP to extrapolate the estimated lower bound on the energy absorption capability of the human biomechanics, and correspondingly, we have designed various nonlinear adaptive stabilizers  \cite{atashzar2017grasp,atashzar2016passivity}. We have shown that by incorporating even the most conservative estimation of EoP into the stabilizer’s design, the performance surpasses the state-of-the-art TDPA (which does not consider the energy absorption of the operator’s biomechanics).\par

\section{Controller Design}

Considering the shortcomings of existing stabilizers and  the benefits brought by biomechanic energy margin, this paper proposes a new stabilizer called Two-Port Biomechanics-aware Passivity-based  Synchronizer and Stabilizer ($ TBPS^2$), which for the first time integrates the energetic behavior of human biomechanics into the two-port stabilizer architecture to enhance the stability while significantly improving position synchronization.\par

When compared with conventional counterparts, the proposed $ TBPS^2$ algorithm can have a better position-domain synchronization between the
leader and follow robot as well as a better stabilization performance without any restriction in the presence of the passivity behavior of the environment and variable unknown communication delay. The remainder of this section will introduce the TBPS\textsuperscript{2} stabilizer architecture and its corresponding passivity criterion, explaining how it works and overcomes the abovementioned issue. \par

\begin{figure}[htbp]
\centerline{\includegraphics[width=0.5\textwidth]{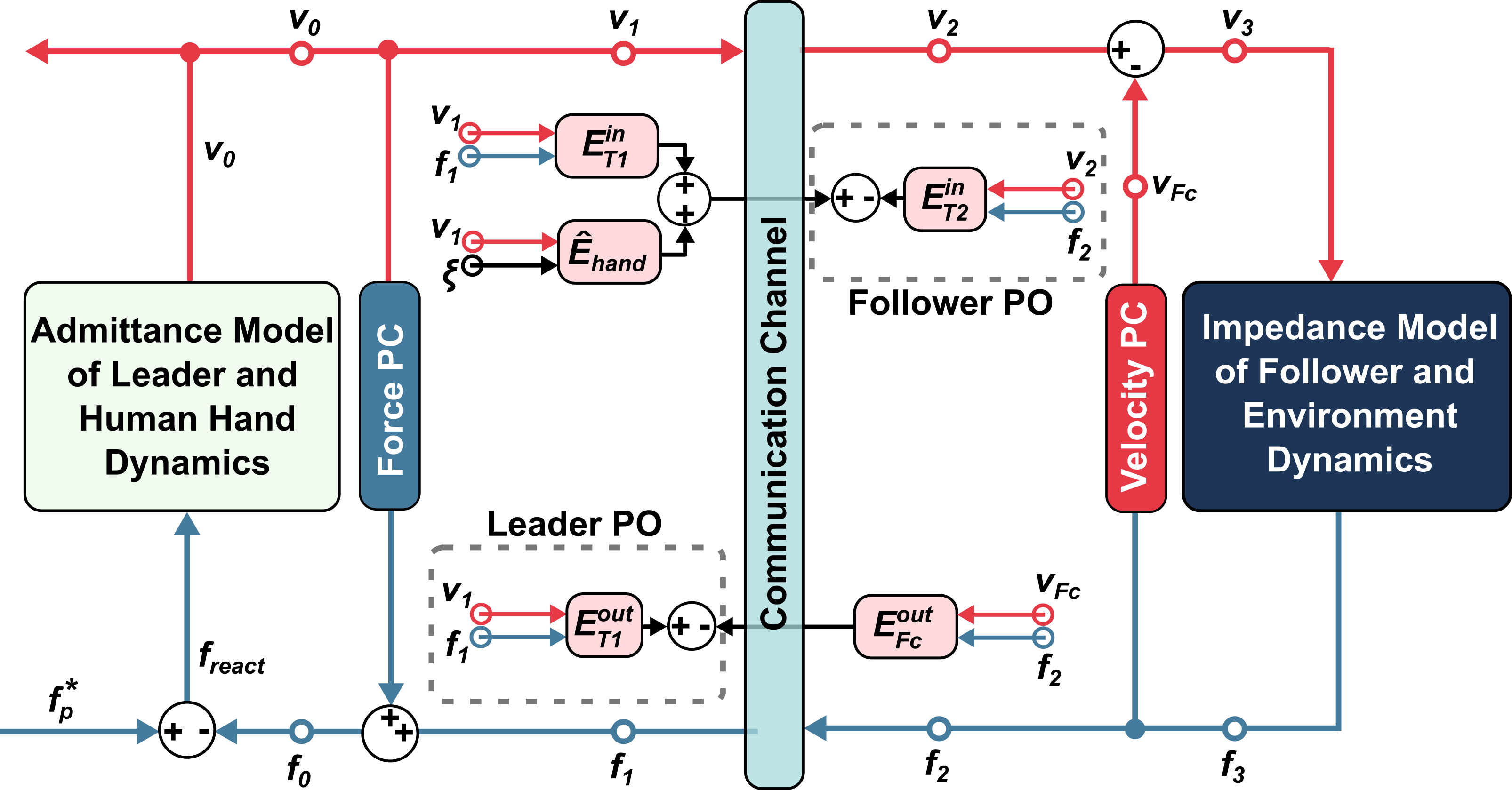}}
\caption{TBPS\textsuperscript{2} Stabilizer architecture.}
\label{fig3}
\end{figure}
The structure of  $ TBPS^2$ contains a force passivity controller on the leader side and a velocity passivity controller on the follower side (Fig. \ref{fig3}), and their corresponding novel designs of PO on each side of the communication, which observe the energy flow. In this work, the estimated EoP and the corresponding biomechanic energy margin that exists in the leader side operator’s limb ($\hat{E}_{hand}$)  are integrated into the stabilizer architecture. The other components of energy flow existing in the system will be described in the following section. \par

In order to design the proposed stabilizer, we consider the velocity decomposition and force equality from Fig. \ref{fig3}:

\begin{equation}\label{v3=v2-vfc}
v_3 = v_2 -v_{Fc},
\end{equation}
\begin{equation}\label{f3=f2}
f_3 = f_2,
\end{equation}
{where} $v_{Fc}$ is the velocity generated by the follower side PC. Accordingly, we can reformulate the passivity criterion (12) based on environmental force ($f_3$) and velocities ($v_3$):

\begin{equation}
\begin{split}\label{exp:A}
A &= \Delta T_L\sum_{i=0}^{k}{[(f_1v_1)^{in}(k) + (f_1v_1)^{out}(k)]} \\
&\quad - \Delta T_F\sum_{i=0}^{k}[(f_2v_2)^{in}(k) - (f_2v_2)^{out}(k) \\
&\quad + (f_3v_3)^{in}(k) + (f_3v_3)^{out}(k)]
\end{split}
\end{equation}

\begin{equation}
\begin{split}\label{exp:B}
B &= \Delta T_L\sum_{i=0}^{k}{[(f_1v_1)^{in}(k) + (f_1v_1)^{out}(k)]} \\
&\quad - \Delta T_F\sum_{i=0}^{k}[(f_2v_2)^{in}(k) \\
&\quad + (f_3v_3)^{in}(k) - (f_2v_{Fc})^{out}(k)]
\end{split}
\end{equation}

\begin{equation}\label{inequality}
A = B \geq 0
\end{equation}

In this equation, the leader and follower may have different sampling rates of `$\Delta_L$' and `$\Delta_F$' respectively.

 We can also decouple  \eqref{18} to separate the overall energy from the energy observed by the leader side PO and the follower side PO separately.
\begin{equation}\label{20}
E^{L}(k)  = \Delta T_L\sum_{i=0}^{k}[(f_1v_1)^{out}(k) - (f_2v_{Fc})^{out}(k)],
\end{equation}
\begin{equation}\label{21}
E^{F}(k)  = \Delta T_F\sum_{i=0}^{k}[(f_1v_1)^{in}(k) + f_3(k)v_3(k))^{in} - (f_2v_{2})^{in}(k)].
\end{equation}
The theoretical energy observed on the leader side (left) is $E^{L}(k)$, and the theoretical energy on the follower side (right) is $E^{F}(k)$. After decoupling, we can rewrite the passivity condition as: 

\begin{equation}\label{22}
 E_{stab}(t)\geq 0, \quad if \quad  E^{L}(k) \geq 0 \quad and \quad E^{F}(k) \geq 0.
\end{equation}
This is a conservative postulate since it requires $E^L(t)$ and $E^F(t)$ both to be greater than zero to satisfy $E_{stab}(t) \geq $0.

However, in reality, due to the communication delay, \eqref{20} and \eqref{21} are not observable since they both depend on the time-shifted energy value on the opposite side of the communication channel\cite{RyuPro2007}. In this case, the observable version of \eqref{20} and \eqref{21} are defined as:

\begin{equation}\label{22}
E_{obs}^{L}(k)  = \Delta T_L\sum_{i=0}^{k}(f_1v_1)^{out}(k) -  \Delta T_F\sum_{i=0}^{k}(f_2v_{Fc})^{out}(k-K_2 ),
\end{equation}
\begin{equation}\label{23}
E_{obs}^{F}(k)  = \Delta T_L\sum_{i=0}^{k}(f_1v_1)^{in}(k-K_1 )  - \Delta T_F\sum_{i=0}^{k}(f_2v_{2})^{in}(k).
\end{equation}
Both $E_{obs}^{L}(k)$ and $E_{obs}^{F}(k)$ are the actual observable versions of energy that can be observed by the leader PO and follower PO, respectively. $K_1$ and $K_2$ represent the discrete-time domain variable delays which may be any non-negative number of time-steps at discrete time-step `k'.

The summation energy terms  $\left | \Delta T\sum_{i=0}^{k}(f_2v_{Fc})^{out}(k)\right |$ and $\Delta T\sum_{i=0}^{k}(f_1v_1)^{in}(k)$ are monotonically increasing, such that: 
\begin{equation}
    \Delta T_L\sum_{i=0}^{k}(f_1v_1)^{in}(k-K_1 ) \leq \Delta T_L\sum_{i=0}^{k}(f_1v_1)^{in}(k),
\end{equation}
\begin{equation}
\left | \Delta T_F\sum_{i=0}^{k}(f_2v_{Fc})^{out}(k-K_2 )  \right | \leq\left |  \Delta T_F\sum_{i=0}^{k}(f_2v_{Fc})^{out}(k) \right |.
\end{equation}
Therefore, the actual (with communication delay) observable version of energy \eqref{22}-\eqref{23}  is a more conservative condition than theoretical (ideal no-delay communication) \eqref{20}-\eqref{21}, to justify the passivity criterion in the stabilizer. This has been mathematically analyzed for any type of non-negative communication delay in \cite{ryu2010passive,RyuPro2007}. In this case, the passivity criterion is:

\begin{equation}\label{27}
 E_{stab}(k)\geq 0, \quad if \quad  E_{obs}^{L}(k) \geq 0 \quad and \quad E_{obs}^{F}(k) \geq 0.
\end{equation}
As mentioned before, integrating the biomechanic energy margin into the two-port stabilizer architecture is another innovation of this paper. As a result, given the lower bound of the operator hand's EoP $\xi _{L-relax}$, we know that at time $t$, the minimum  energy the hand can absorb is:

\begin{equation} \label{Ehand}
\hat{E}_{hand}(k) = \Delta T_L\sum_{i=0}^{k}\xi _{L-relax}v_{0}(k)^{T}v_{0}(k),
\end{equation}
where $\hat{E}_{hand}(k)$ is the minimum  energy absorption capability of the operator’s hand in real-time. Note: since the lower-bound $\xi_{L-relax}$ is measured offline, the real-time energy absorption capability calculated in \eqref{Ehand} is a lower-bound estimate that relies only on the real-time velocity measurement. After integrating this new element, the theoretical passivity condition of the overall TBPS\textsuperscript{2}, including the operator’s biomechanics and the remaining stabilizer architecture is:

\begin{equation}\label{29}
\hat{E}_{hand}(k)+ E_{stab}(k) \geq 0.
\end{equation}
However, the $\hat{E}_{hand}(k)$ is measured on the leader side and utilized on the follower side for position drift compensation, which will suffer the communication delay in practical. Therefore, combined with \eqref{27}, the observable version of the TBPS\textsuperscript{2} passivity condition of \eqref{29} is

\begin{equation}
\hat{E}_{hand}(k-K_1)+ E_{obs}^{L}(k) + E_{obs}^{F}(k) \geq 0,
\end{equation}
where $\hat{E}_{hand}(k-K_1)$ is the observed version of $\hat{E}_{hand}(k)$ on follower side.

\begin{figure}[htbp]
\centerline{\includegraphics[width=0.48\textwidth]{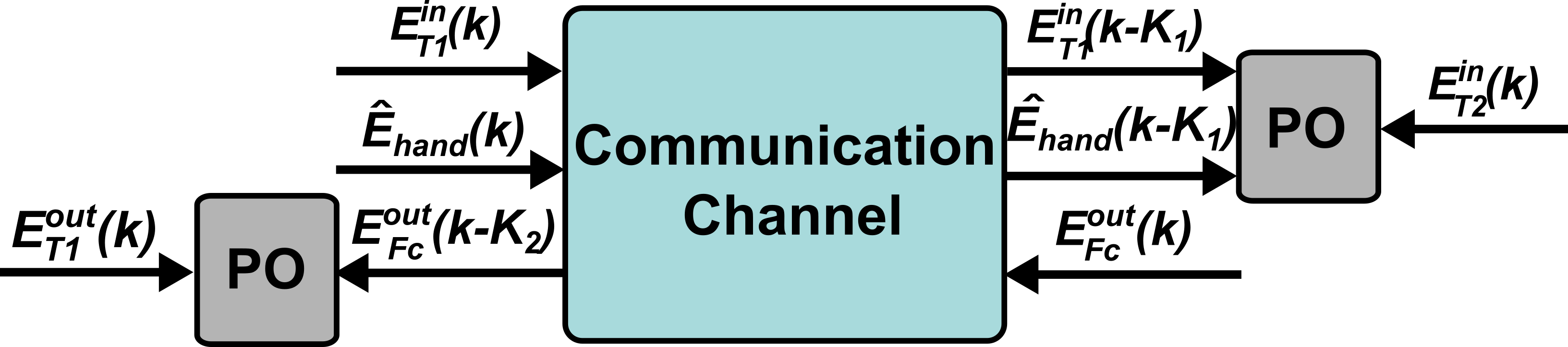}}
\caption{Energy flow through the Communication Channel}
\label{fig4}
\end{figure}

To better visualize the observable version of energy flow, Fig. \ref{fig4} shows the overall energy flow that existed in the two-port communication networked system. The leader side PO observed the output energy from observation terminal 1 $E_{T1}^{out}(k)$ and the delayed version of the output energy generated by the follower PC $E_{Fc}^{out}(k-K_2)$. The follower side PO observed the input energy from observation terminal 2  $E_{T2}^{in}(k)$, the delay version of the input energy from observation terminal 1 {$E_{T1}^{in}(k-K_1)$, and the delay version of the energy absorption capability of the operator’s hand.

\subsection{Passivity Controller Algorithms}
In this section,  we will introduce the passivity controller design on both sides of the communication channel. The PC is based on the observed energy information to generate the adaptive damping coefficients when needed to modify the force and velocity to guarantee the system's passivity. In the following, the coefficients are formulated for the discrete-time system, which is more common in the practical case.\par

\textbf{Leader side}:
On the leader side, the total energy that the PO observes at \textit{$k^{th}$} time stamp is equal to 
\begin{equation}
 E_{obs}^{L}[k] + E_{Lc}[k-1] =  E_{T1}^{out}[k] - E_{Fc}^{out}[k-K_2] + E_{Lc}[k-1],
\end{equation}
where $E_{Lc}[k-1]$ is the total energy generated by the leader side PC until the last time stamp. $E_{Fc}^{out}[k-K_2]$ is the total energy generated by the follower side controller and delivered to the leader side after \textit{$K_2$} unit delay and \textit{K} is the current time stamp.

If the observed energy is negative, the passivity controller on the leader side adaptively generates the damping coefficient $\alpha$ to modify the force, thus dissipating the excess energy. In this case, the leader side PC can be designed as:

\begin{equation}
  \alpha[n]=\begin{cases}
    \frac{ \eta [k] = E_{obs}^{L}[k] + E_{Lc}[k-1]}{-\Delta T_Lv_{0}[k]^{2}}, & \text{$\eta [k] < 0$}.\\\\
    0, & \text{otherwise}.
  \end{cases}
\end{equation}
\\
\begin{equation}
\begin{aligned}
    f_0[k] &= f_1[k] + f_{Lc}[k],
\end{aligned}
\end{equation}
\begin{equation}
    f_{Lc}[k]  = \alpha [k]v_0[k],\\
\end{equation}

\begin{align}
        E _{Lc}[k-1] &= \Delta T_L\sum_{i=0}^{k} f_{Lc}[k-1]v_0[k-1].
\end{align}

\textbf{Follower side}:
On the follower side, the total energy that the PO observe at \textit{k} time stamp is equal to:
\begin{equation}
\begin{aligned}
&E_{obs}^{F}[k]+ \hat{E}_{hand}[k-K_1]+ E_{Fc}[k-1] = \\
&E_{T1}^{in}[k-K_1] - E_{T2}^{in}[k]  + \hat{E}_{hand}[k-K_1] + E_{Fc}[k-1],
\end{aligned}
\end{equation}
where $E_{Fc}[k-1]$ is the total energy generated by the follower side PC until the last stamp, and $\hat{E}_{hand}[k-K_1]$ is the observable version of the biomechanical energy margin that exists in the operator limb.

The PC will generate the coefficient $\beta$ to either (i) remove non-passivity or (ii) correct position drift by adjusting the velocity based on the energy information. If the PO detects that the current energy is negative, it means that the system is non-passive and potentially unstable. In that case, the PC will generate the positive damping parameter $\beta$ to modify the velocity and then dissipate the extra energy, guaranteeing the system’s stability. On the other hand, if the PO detects that the current energy is positive, it means that the system is stable and there exists an extra energy margin (margin of stability) that can be used to compensate for the position drift between two robots. In that case, the PC generates the damping parameter $\beta$ (sign depends on the position compensation direction) to adjust the velocity and mitigate the position error without violating the passivity criterion. 

In order to compensate for the position drift, first, we define the real-time position drift between two robots. Mathematically, the modified velocity applied to the follower robot is:

\begin{equation}
    v_{3}(k) = v_{2}(k) - \beta f_{2}(k),
\end{equation}
where $v_{2}(k)$ is the desired velocity transmitted from the leader robot side, $v_{3}(k)$ is the modified velocity that actually applied to the follower robot.

We can integrate $v_{3}(k)$ to get the actual position of the follower robot $x_{3}(k)$:

\begin{equation}
        x_{3}(k) = \Delta T_F\sum_{i=0}^{k}v_{3}(k).
\end{equation}
{Similarly, the leader side robot position $x_{2}(k)$ is:}

\begin{equation}
        x_{2}(k) = \Delta T_F\sum_{i=0}^{k}v_{2}(k).
\end{equation}
{The difference between  $x_{2}$ and $ x_{3}$ is the position drift between the leader and follower robot, which is $\Delta x(t)$:}

\begin{equation}
    \Delta x(k) = \Delta T_F\sum_{i=0}^{k}v_{2}(k) - \Delta T_F\sum_{i=0}^{k}v_{3}(k).
\end{equation}
{Moreover, if the follower side total observed energy is positive, the maximum position drift $\Delta x_{max}(k)$ that can be compensated without violating the passivity criterion is:}

\begin{equation}
    \Delta x_{max}(k) = \frac{E_{obs}^{F}[k]+ \hat{E}_{hand}[k-K_1]+ E_{Fc}[k-1]}{f_{3}[k]}.
\end{equation}
As a result, the follower side passivity controller can be designed as:

\begin{equation}
  \beta[n]:\begin{cases}
     \frac{\eta [k] = E_{obs}^{F}[k]+ \hat{E}_{hand}[k-K_1]+ E_{Fc}[k-1]}{-\Delta T_Ff_{3}[k]^{2}},  & \text{if $\eta [k]< 0$}.\\\\
    \frac{min(\left | \Delta x\right |,\left | \Delta x_{max} \right |)}{Sgn(\Delta x)f_{3}[k]\Delta T_F}, & \text{otherwise}.
  \end{cases}
\end{equation}

\begin{equation}
    v_{3}(k) = v_{2}(k) - \beta(k) f_{3}(k),
\end{equation}
\begin{equation}
    v_{Fc}(k) = \beta(k) f_{3}(k),
\end{equation}

\begin{align}
E_{Fc}[k-1] &= \Delta T_F\sum_{i=0}^{k} v_{Fc}[k-1]f_3[k-1],
\end{align}

where \textit{Sgn} is the sign function used to determine the position compensation direction. {The dimensions of $\alpha[n]$ and $\beta[n]$ are defined by the leader and follower robot's degree of freedom (DoF) and does not affect the stabilizer's performance.}\par
The design of $\beta$, which is a key aspect of this paper, follows a structure similar to that found in \cite{artigas2010position}. However, the ability to cancel position drift is significantly improved. 
To be more specific, by considering the energy absorption capabilities of human biomechanics, the stabilizer is granted an additional margin that enables the system to remain passive in terms of position energy, which reduces controller activation and subsequently minimizes velocity modification and position drift. Furthermore, suppose the system becomes non-passive and a certain amount of position drift is present in the teleoperation system. In that case, the added energy absorption capability of human biomechanics allows the controller to compensate for a larger maximum position drift distance than if this factor were not taken into account.

\section{Simulation results}

\begin{table}[H]
\centering
\caption{Simulation Parameter}
\begin{tabular}{|l|l|} 
\hline
\textbf{Operator Model:}     & $Z_h^{-1}$ = $\frac{s}{0.5s^2+50s+10} $         \\ 
\hline
\textbf{Exogenous Force:}    & $f^*$= $100(sin(0.5t)+cos(5t))$ N  \\ 
\hline
\textbf{Excess of Passivity:} & 25 Ns/m                   \\
\hline
\textbf{Environment Dynamics:} & $Z_e$ = $B_e + 80s$                \\
\hline
\textbf{Communication Delay:} & $t_d$ = $(Delay + 0.1Delay*cos(t)$) s                \\
\hline
\end{tabular}
\end{table}

To verify the efficacy of the proposed stabilizer, systematic grid simulations comprised of 256 sets of parameters are tested. The proposed TBPS\textsuperscript{2} is tested against two different types of controllers: (1) state-of-the-art controllers with both velocity and force modification, and (2) state-of-the-art controllers with only force modification. Group (1) controllers contain: chattering free TDPA (TDPA-CF) \cite{tdpa-cf}, TDPA with position synchronization and force/torque smoothening (TDPA-PS) \cite{tdpa-ps}, and TDPA with position synchronization through energy injection (TDPA-Eng) \cite{tdpa-eng}. For this group, the stabilization effort of the PC at each port using the Pearson Correlation Coefficient, and the average position drift are evaluated. Group (2) contains the single-sided TDPA using an energy storage method (TDPA-Rad) \cite{tdpa-rad}. For this group, only the stabilization effort of the PC at the leader side using the Pearson Correlation Coefficient is evaluated. The rest of the results are organized into the following subsections: subsection `A' are the grid simulation results comparing the TBPS\textsuperscript{2} to group (1) and (2) in a resistive environment (which impedes the motion of the leader operator), subsection `B' is a verification of the stabilizers to remain passive in an assistive environment (which enhances the motion of the leader operator), and subsection `C' are the results of the assistive environment grid simulation.\par


Table 1 shows the simulation settings. We denote the modeled admittance of the operator’s hand on the leader side as $Z_{h}^{-1}$. To implement the proposed controller, we utilized a lower bound EoP in order to simulate a realistic condition when the exact estimate of EoP may not be accessible (for details in this regard, please check our previous works \cite{atashzar2017grasp,sfaRALhip}). $f^*$ is the exogenous force input by the operator on the leader side, and $B_e$ is the environmental damping coefficient. The communication delay is made to be a variable delay.


\subsection{Resistive Environment:}

The resistive environment is modeled as a spring-damper system. For the grid simulation, the environmental damping $B_e$ is varied from 0 $Ns/m$ to 90 $Ns/m$ with 16 equal step-sizes resulting in 16 different $B_e$ values. Likewise, the one way communication delay magnitude `$Delay$' is varied from 0 $s$ to 0.3 $s$ with 16 equal step-sizes. This results in a total of 256 simulation trials for each stabilizer.

\subsubsection{ Spearman Correlation for Velocity Tracking}

In order to compare the desired velocity at the follower side $v_2$ (which is the velocity sent over the communication network) and the modified velocity $v_3$ after the controller at the follower side, we measure the Spearman Correlation Coefficient.\par

\textbf{Group 1:} The results for the Velocity Spearman Correlation Coefficient are provided in Fig. \ref{VRR}.


\begin{figure}[http]
\centerline{\includegraphics[width=0.5\textwidth]{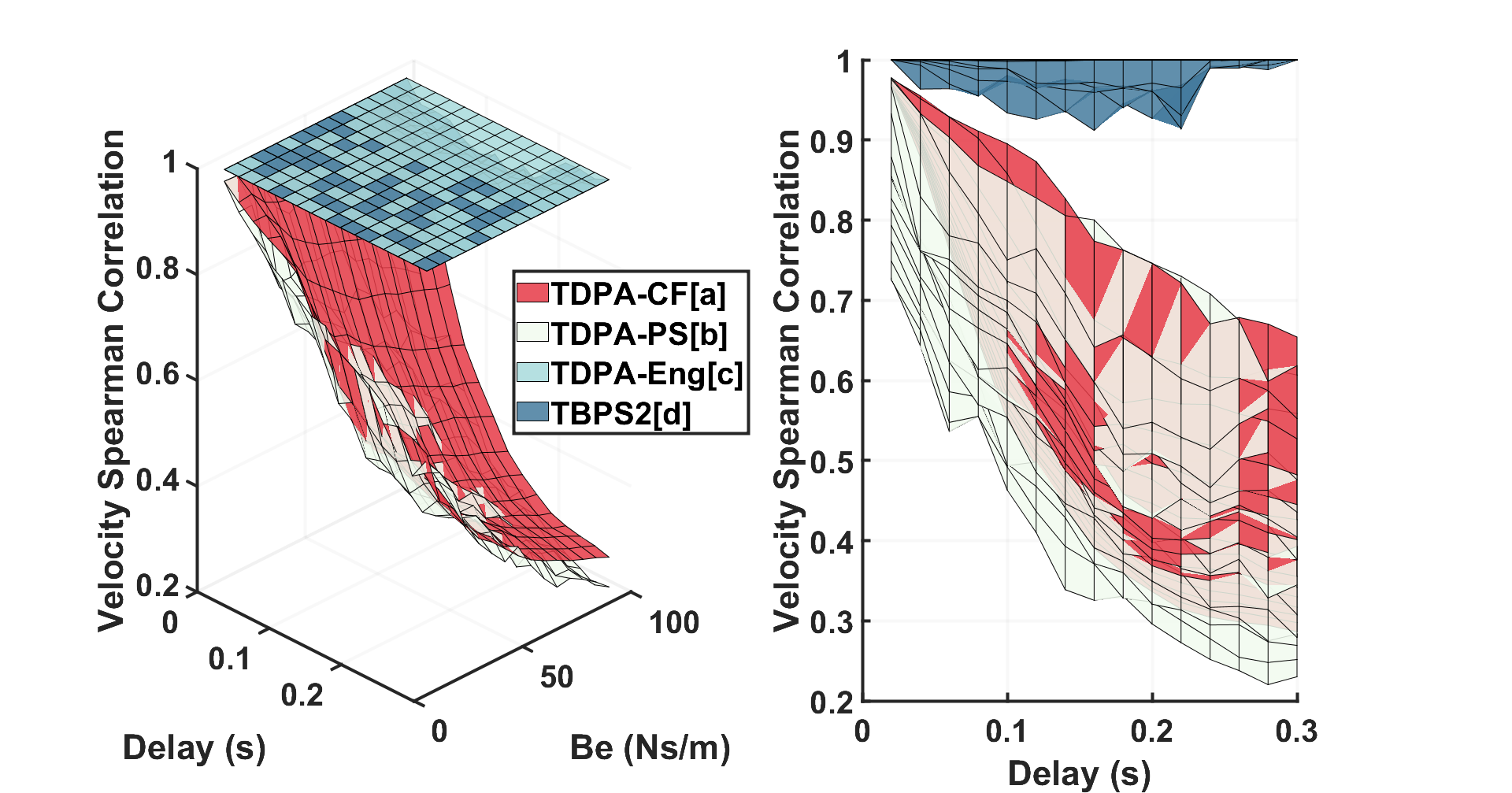}}
\caption{The Surface plot of Spearman Correlation between $v_2$ and $v_3$ as a function of delay and $B_e$ for TBPS\textsuperscript{2} and Group 1 stabilizers (Left). The Surface plot is rotated to focus on the Spearman Correlation as a function of delay only (Right).}
\label{VRR}
\end{figure}

In this figure, the Spearman Correlation between $v_2$ and $v_3$ is plotted for TBPS\textsuperscript{2} and the Group 1 stabilizers. Spearman Correlation captures the monotonic nonlinear and non-parametric relationships between two variables. A Spearman Correlation Coefficient close to 1 indicates that $v_2$ and $v_3$ have a strong monotonic relationship maintained by the controller and a low Spearman Correlation shows a low-quality velocity tracking. 
As we can see, the Spearman's Correlation of the TBPS\textsuperscript{2} is close to unity and superior to the Spearman Correlation of TDPA-PS and TDPA-CF over all the Delay-$B_e$ grid. The exception is the TDPA-Eng which also shows high Spearman Correlation Coefficients.



\textbf{Group 2:} Since the velocity has no modification for this group, the RMSE is 0 and the Velocity Correlation Coefficient is 1 in all parameter sets.

\subsubsection{Position Drift}

To evaluate the stabilizer's position synchronization performance, the average position drift between the desired and actual motion for the grid simulation is shown in Fig. \ref{drift}.

\begin{figure}[!t]
\centerline{\includegraphics[width=0.5\textwidth]{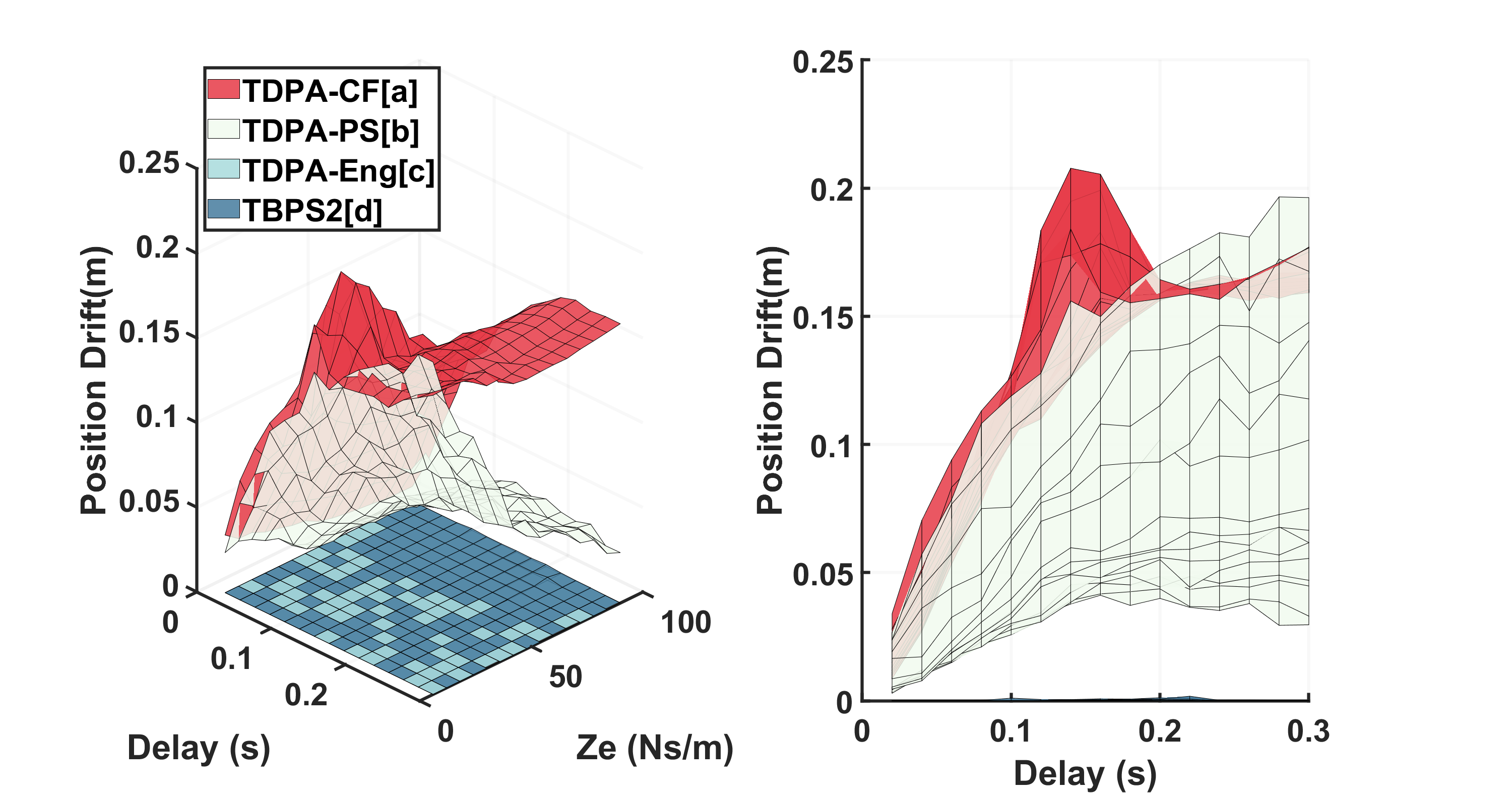}} 
\caption{The Surface plot of the position drift between leader side and follower side as a function of delay and $Z_e$ for TBPS\textsuperscript{2} and Group 1 stabilizers (Left). The 3-D plot is rotated to focus on the position drift as a function of delay only (Right).}
\label{drift}
\end{figure}

\begin{figure}[!t]
\centerline{\includegraphics[width=0.3\textwidth]{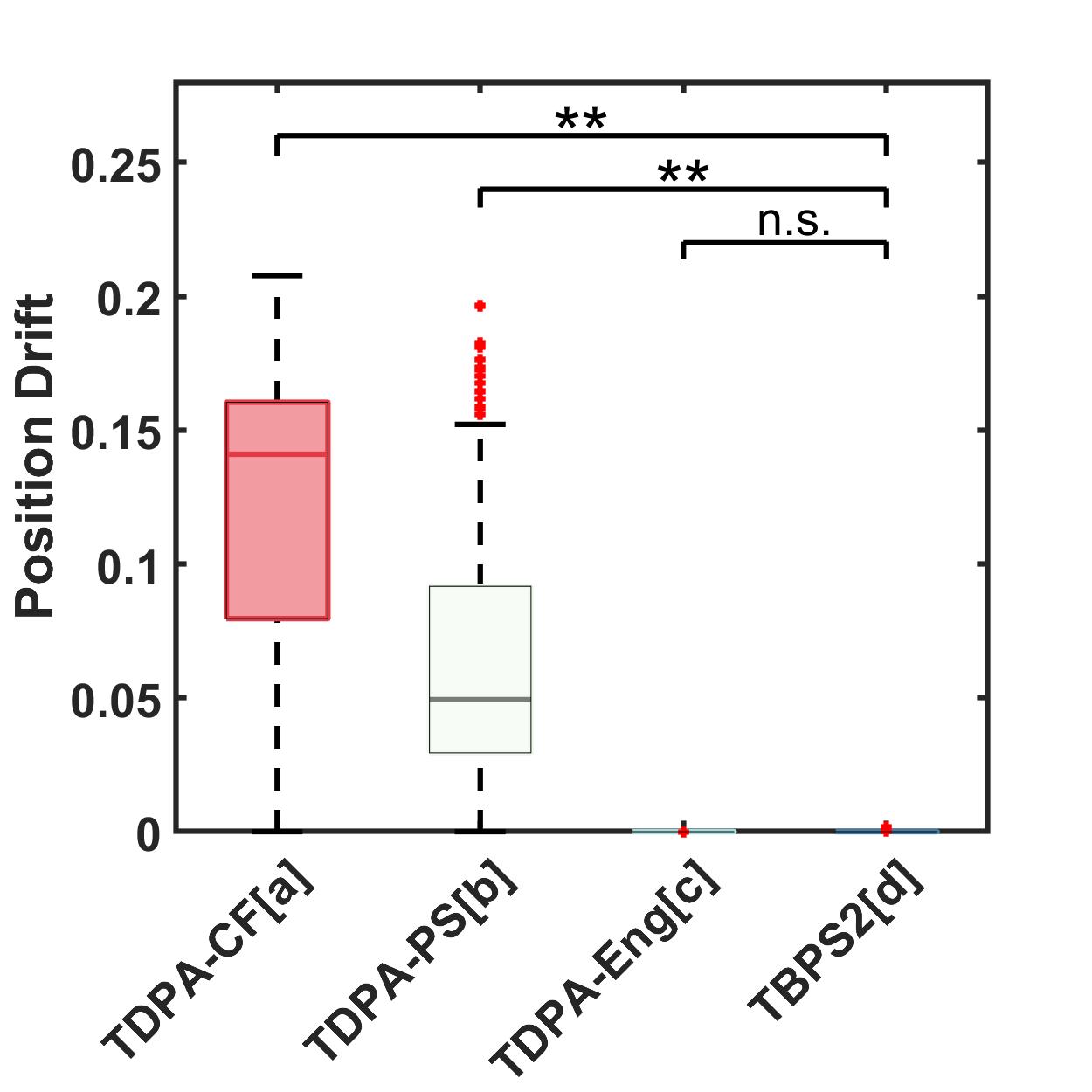}}
\caption{The resulting box plot distribution of the average position drifts for Group 1 stabilizers.}
\label{drift_box}
\end{figure}

Fig. \ref{drift} denotes the position drift between the leader and the follower robots averaged over the duration of the trial for the mentioned Delay-$B_e$ grid simulation. A value close to 0 indicates that the two robot movements are synchronized in the “position domain” with minimum position drift. As can be seen in the figure, the position drift for TBPS\textsuperscript{2} is generally closer to 0 for all simulated parameters. This is also the case for the TDPA-Eng which presented peak velocity tracking in the form of the Velocity Correlation Coefficient. However, the TDPA-CF and TDPA-PS have position drifts that increase linearly as the delay increases. As a result, we concluded that the TBPS\textsuperscript{2} and TDPA-Eng outperform the state-of-the-art TDPA-PS in terms of position synchronization. The TBPS\textsuperscript{2} achieves this by utilizing a biomechanical energy margin as an additional source of passivity.

The average position drifts are separated into their own distribution by stabilizer resulting in a total of `$n=256$' values for each distribution. The Kolmogorov-Smirnov normality tests showed that the distributions were non-normal \cite{stat2}. Therefore, the Wilcoxon signed-rank test was used to determine whether the distributions of two stabilizer groups had statistical significance in their medians using a significance level of 5\% \cite{stat1}. The resulting box plot and significance levels are shown in Fig. \ref{drift_box}. These results reflect the observations in Fig. \ref{drift}. The position synchronization for the proposed TBPS\textsuperscript{2} and the TDPA-Eng are similar and show excellent synchronization (drift is nearly 0). Likewise, the proposed TBPS\textsuperscript{2} has a significantly lower position drift compared to the TDPA-CF and TDPA-PS.

\subsubsection{Spearman Correlation for Force Tracking}

Here, we compare the desired force at the leader side $f_1$ (which is the force sent over the communication network to the leader side)  and the modified force $f_0$ after the controller at the leader side using the Spearman Correlation Coefficient.\par

\textbf{Group 1:} The results for the Force Spearman Correlation Coefficient are provided Fig. \ref{FRR1}.

\begin{figure}[!t]
\centerline{\includegraphics[width=0.5\textwidth]{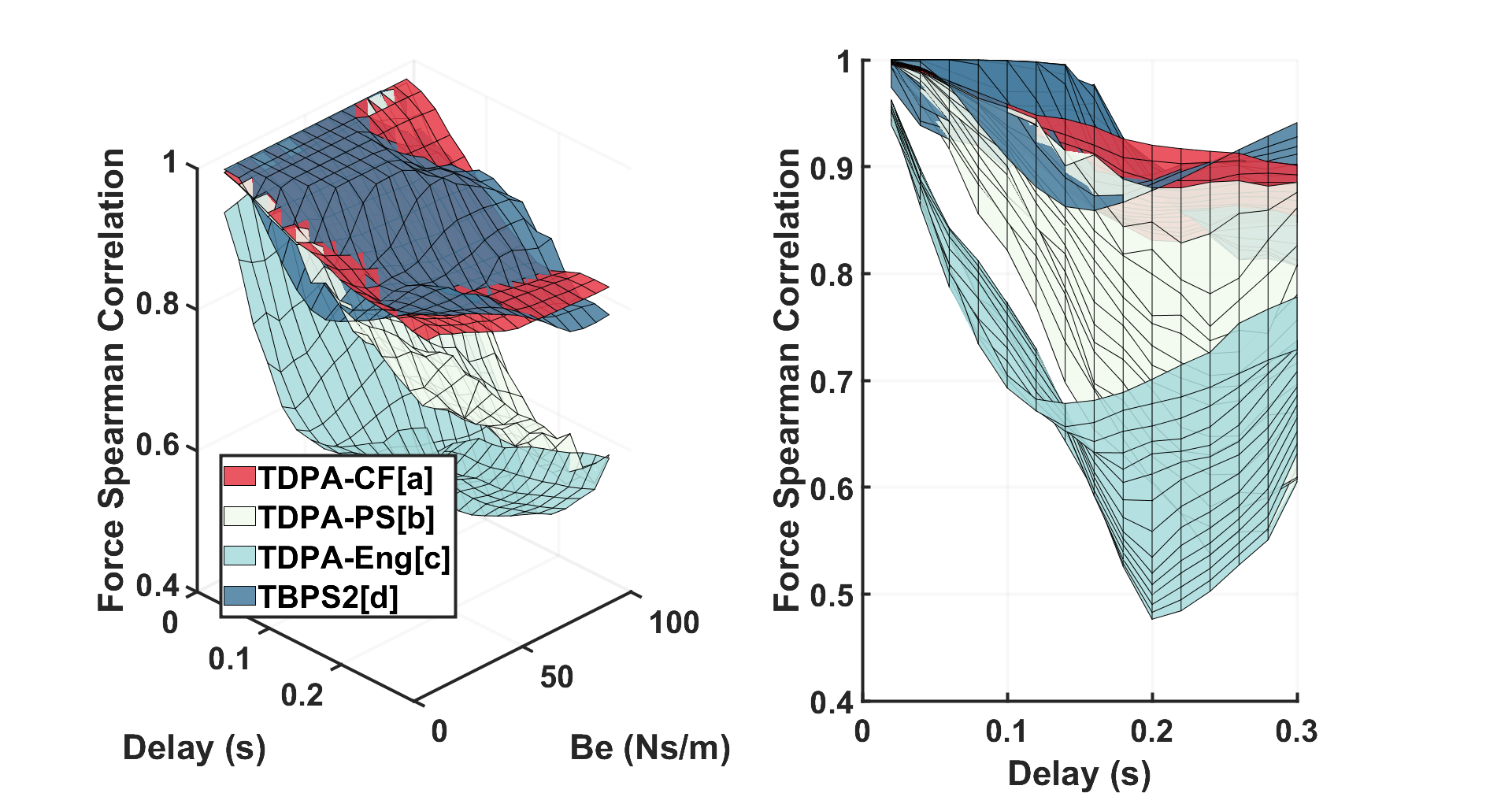}}
\caption{The Surface plot of Spearman Correlation between $f_1$ and $f_0$ as a function of delay and $B_e$ for TBPS\textsuperscript{2} and Group 1 stabilizers (Left). {The side view of the Surface plot shows the Spearman Correlation as a function of delay (Right).}}
\label{FRR1}
\end{figure}

\begin{figure}[!t]
\centerline{\includegraphics[width=0.3\textwidth]{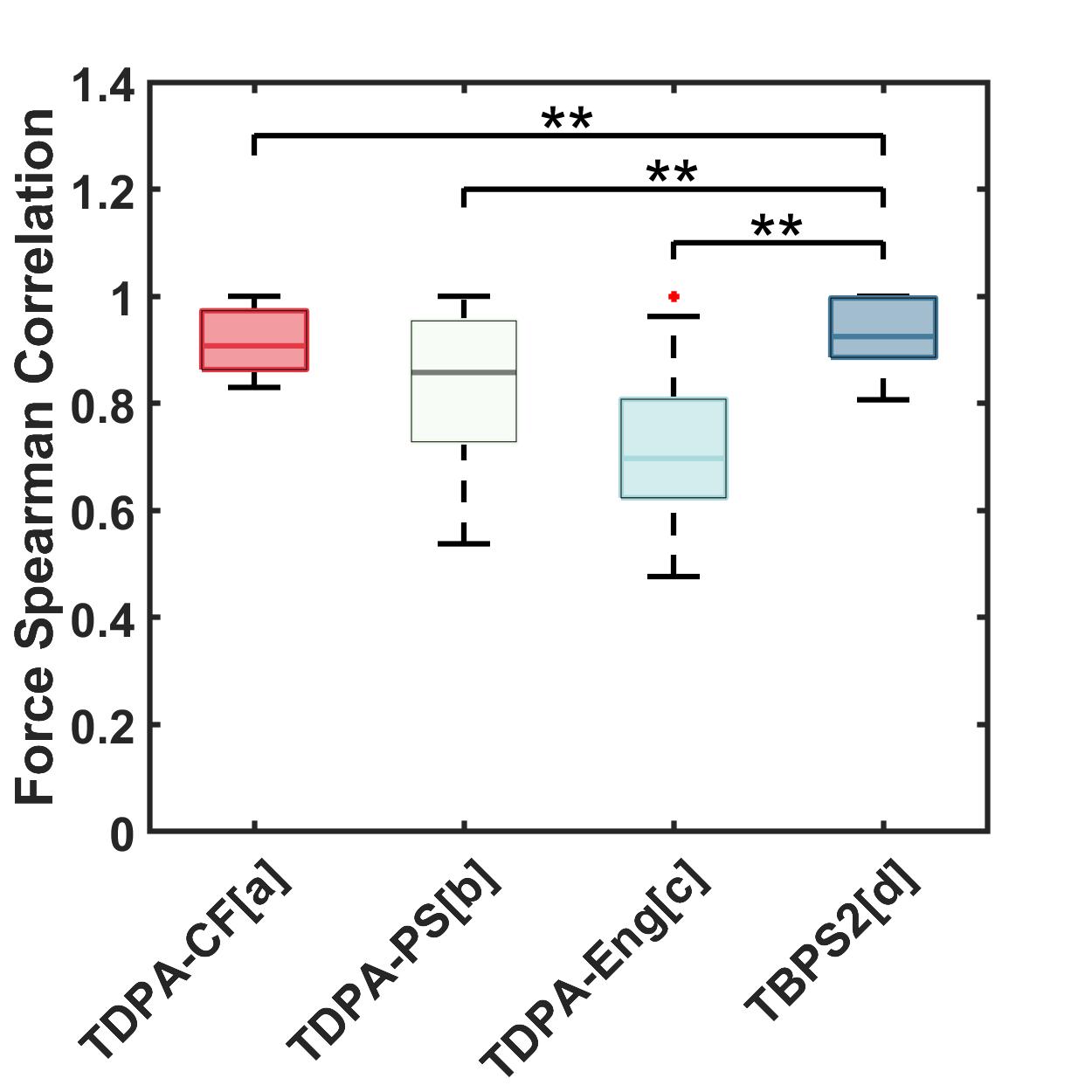}}
\caption{The resulting box plot distribution of the Force Correlation Coefficients for Group 1 stabilizers.}
\label{FRR_box}
\end{figure}

Similarly to the position drift results, the Force Spearman Correlation Coefficient results are separated into four distributions, and the Wilcoxon signed-rank test is performed (distributions are non-normal). The resulting box plots are shown in Fig. \ref{FRR_box}. The Force Spearman Correlation Coefficient of the proposed TBPS\textsuperscript{2} is higher than that of the TDPA-PS and TDPA-Eng. In fact, these values are lowest for the TDPA-Eng. Often, stabilizers which provide good velocity tracking perform poorly on the force tracking as a trade-off. However, for the proposed stabilizer, it can be seen to have good force and velocity tracking as a result of its use of the human biomechanics capability to absorb additional energy.

\textbf{Group 2:} In regards to this, we also evaluate the Force Spearman Correlation Coefficient with the TBPS-Rad which only provides force modification. The surface plot results of the grid simulation are provided in Fig. \ref{FRR2}. For the TBPS-Rad, the entire controller effort is placed on the force modification on the leader side; as a trade-off, it has very good velocity tracking. Since the proposed TBPS\textsuperscript{2} has placed the stabilization burden on both side controller, it has a superior force tracking performance than the TDPA-Rad.

\begin{figure}[!t]
\centerline{\includegraphics[width=0.5\textwidth]{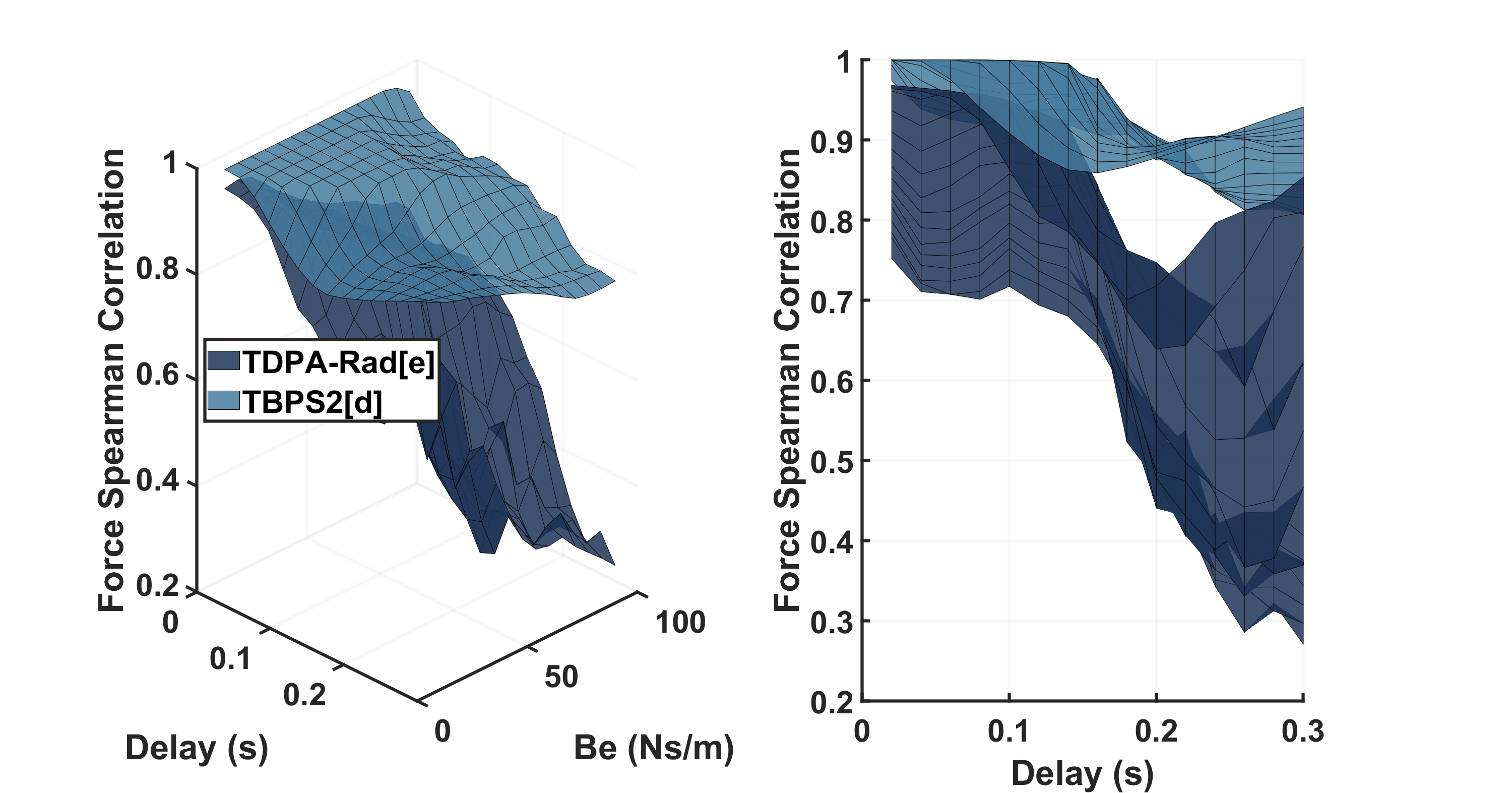}}
\caption{The Surface plot of Spearman Correlation between $f_1$ and $f_0$ as a function of delay and $B_e$ for TBPS\textsuperscript{2} and Group 2 stabilizers (Left). The sideview of the Surface plot showing the Spearman Correlation as a function of delay (Right).}
\label{FRR2}
\end{figure}




\subsubsection{Demonstration of Stabilizers in a Case Simulation}

\begin{figure}[http]
\centerline{\includegraphics[width=0.5\textwidth]{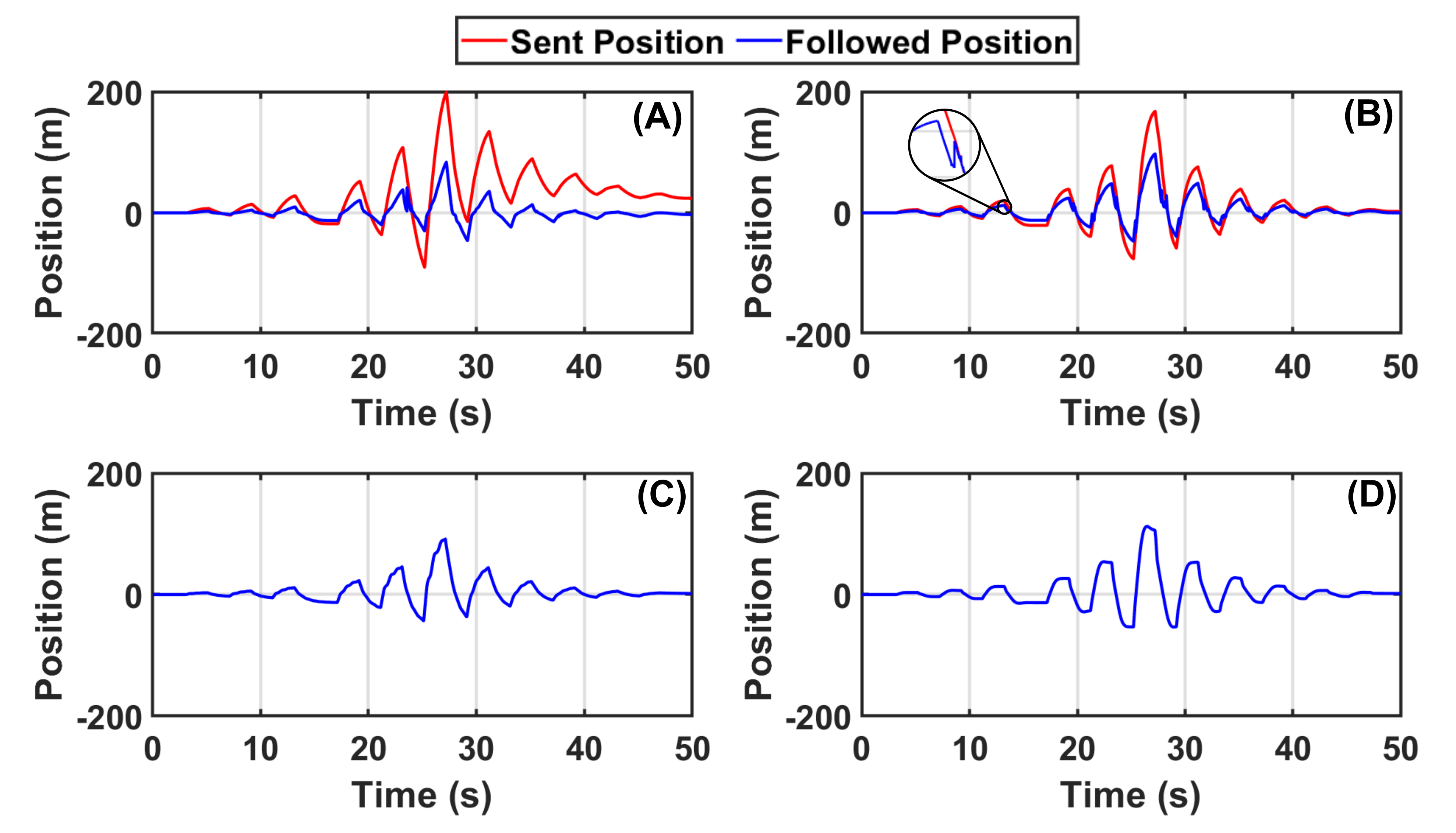}} 
\caption{
Case simulation from the resistive environment. (A) The position tracking of the system under TDPA-CF. (B) The position tracking of the system under TDPA-PS. (C) The position tracking of the system under TDPA-Eng. (D) The position tracking of the system under TBPS\textsuperscript{2}. }
\label{fig9}
\end{figure}

In order to have a close-up demonstration of the performance of the above-mentioned Group 1 controllers, a 50-second case simulation is given using a simulated parameter set (delay = 200 ms, $B_e$ = 12 Ns/m), and an aperiodic square wave signal is used as the exogenous force. As can be seen in Fig. \ref{fig9}, the TDPA-PS (Fig. \ref{fig9} (B)) at some periods exhibits more accurate position synchronization than the TDPA-CF (Fig. \ref{fig9} (A)). This is because this stabilizer utilizes the positive energy that is 'trapped' in the communication network, to partially address the position drift problem. However, due to the limited availability of trapped energy, the position synchronization is also limited. In addition, it should also be noted that in practice, due to the stochastic and unpredictable behavior of communication delay, the synchronization performance can vary and be even lower. The position synchronization between the leader and follower robots under the TDPA-Eng and proposed TBPS\textsuperscript{2}  (Fig. \ref{fig9} (C) and (D), respectively) is significantly more accurate than under TDPA-PS and TPDA-CF, which highlights the excellent position tracking of TBPS\textsuperscript{2} in comparison with other state-of-the-art stabilizers. This phenomenon indicates that after integrating the biomechanical energy margin into the design of TBPS\textsuperscript{2}, the stabilizer has an extra positive energy package to compensate for the existing position drift and thus secure a very high position synchronization performance.

\subsection{Assistive Environment Verification:}

Before starting the systematic grid-based large-scale simulation in the assistive environment, a verification test through impulse response analysis is performed to check whether the Group 1 stabilizers can remain passive in an assistive environment. For this, we simulated the environment and set the damping coefficient to $B_e$ = -100 Ns/m to mimic a highly assistive environment. To investigate the impulse response, the exogenous force ($f_p^*$) was set as an impulse force. The paired velocity-position, and paired power-energy responses over time of all Group 1 stabilizers are investigated with the parameters above and discussed below.

\begin{figure}[http]
\centerline{\includegraphics[width=0.5\textwidth]{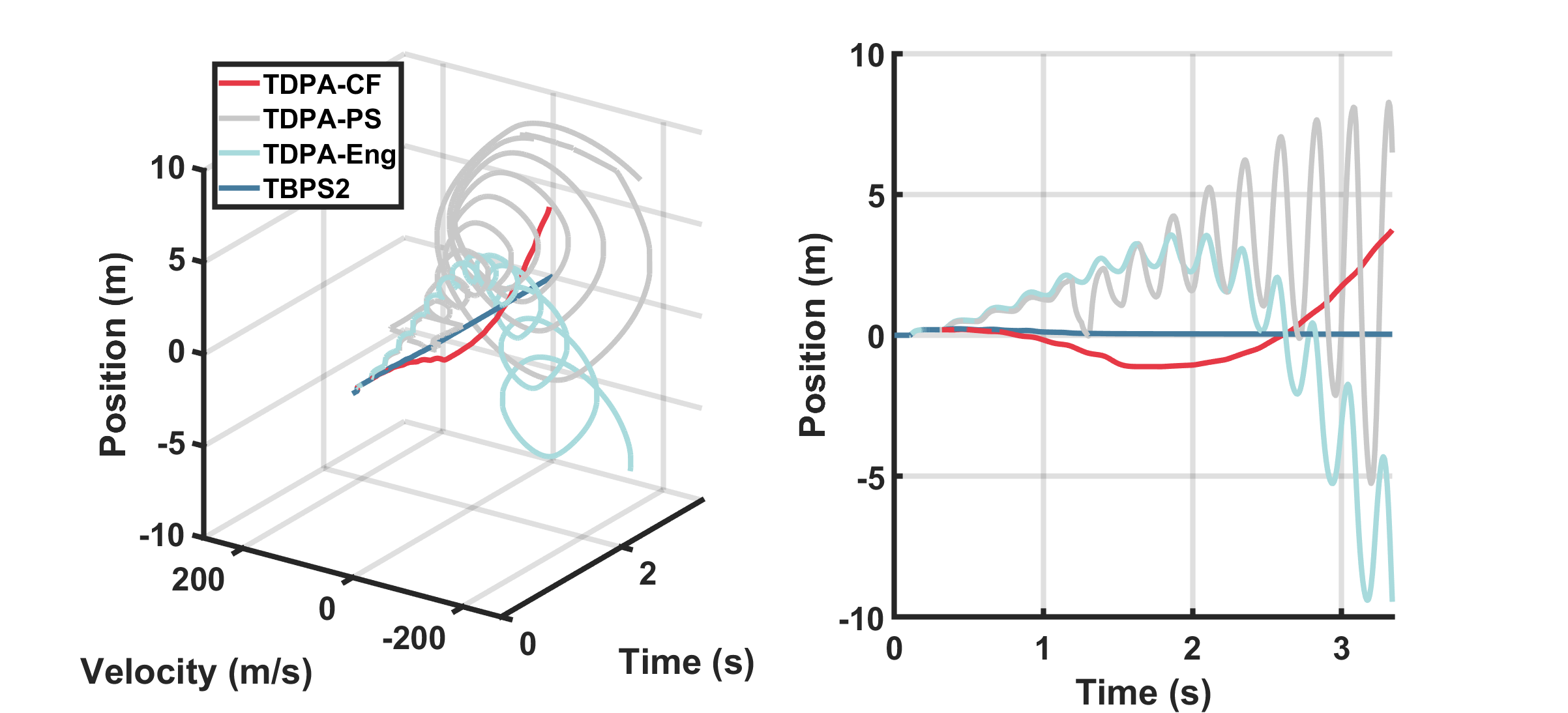}}
\caption{Paired Velocity-Position response of TBPS\textsuperscript{2} and Group 1 Stabilizers under 100ms delay}
\label{impulse_vp}
\end{figure}

\textit{1) Velocity and Position response over time:}\\
Fig. \ref{impulse_vp} shows the stabilizers' velocity and position responses over time. More specifically, the follower side’s position and velocity are shown as a function of time. Each plot includes the stabilizer performances under a simulated 100 ms communication time delay. As can be seen in Fig. \ref{impulse_vp}, after receiving the impulse force, the velocity and position increases and diverges with some oscillation for the Group 1 stabilizers. However, for the TBPS\textsuperscript{2}, the position and velocity increase initially, but eventually converge to zero. These results indicate that the TBPS\textsuperscript{2} controller has the capability to guarantee the system’s stability under the assistive environment even when delay is present. At the same time, these results indicate the other Group 1 stabilizers cannot guarantee the stability and convergence of networked telerobotic systems in an assistive environment.

\begin{figure}[http]
\centerline{\includegraphics[width=0.5\textwidth]{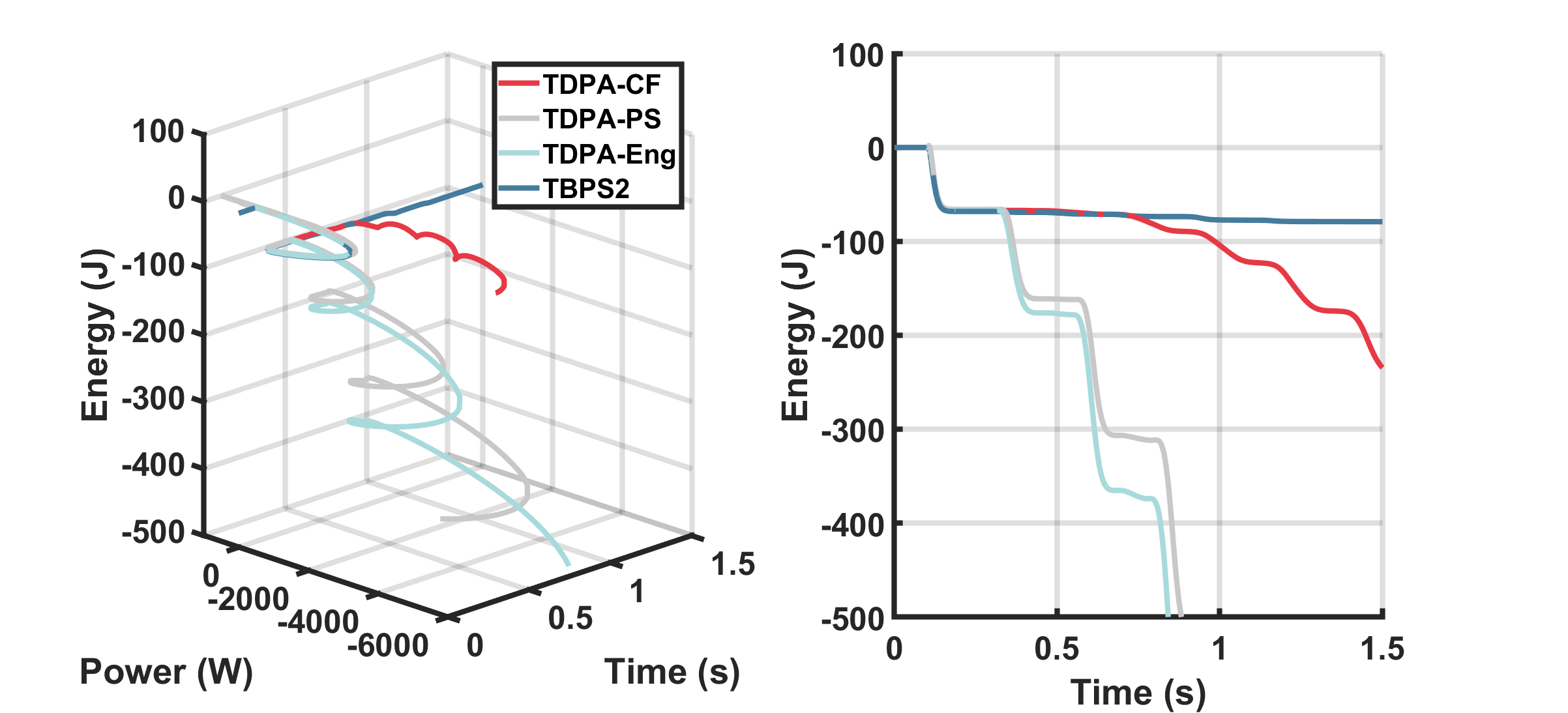}}
\caption{Paired Power-Energy response of TBPS\textsuperscript{2} and Group 1 Stabilizers under 100 ms delay}
\label{impulse_pe}
\end{figure}

\textit{2) Power and Energy response:}\\
The follower side's Power and Energy responses for the Group 1 Stabilizers and TBPS\textsuperscript{2} are shown in Fig. \ref{impulse_pe}. In this regard, the follower side's energy and instantaneous power are shown as a function of time. Each plot includes the controller performance resulting paired power-energy response when provided an impulse.
 As can be seen, after receiving the impulse force, the energy under  TBPS\textsuperscript{2}  increases initially but converges to a constant negative value. At the same time, the power initially increases but converges to 0. However, the other Group 1 stabilizers' power and energy diverge and become more negative. This indicates that the TBPS\textsuperscript{2} controller can restrict the negative energy package growth during the assistive environment can guarantee the system's stability. However, for the other Group 1 Stabilizers, after receiving the impulse force, the energy oscillates and diverges. This indicates that the other Group 1 Stabilizers cannot maintain the system's stability in an assistive environment, { Resistive Environment} which may be a requirement in various practical applications such as telerobotic rehabilitation.

\subsection{Assistive environment: Grid Simulation}

Based on the initial assessment conducted under Section IV.B, the Group 1 stabilizers operate under the assumption of {the} resistive environment, but, become unstable when the environment is assitive. However, the TBPS\textsuperscript{2} stabilizer functions effectively in both the resistive and assistive environments under various communication delay scenarios. To further analyze the behavior of the TBPS\textsuperscript{2} stabilizer under various combinations of the delay and environmental impedance, we conducted a large-scale grid simulation, including 256 trials (with 16 intermediate delay values and 16 intermediate damping coefficients), where the assistive environment damping coefficient $B_e$ ranged from -90 N.s/m to 0 N.s/m. Other environmental parameters are unchanged from the resistive environment simulation.

\begin{figure}[http]
\centerline{\includegraphics[width=0.5\textwidth]{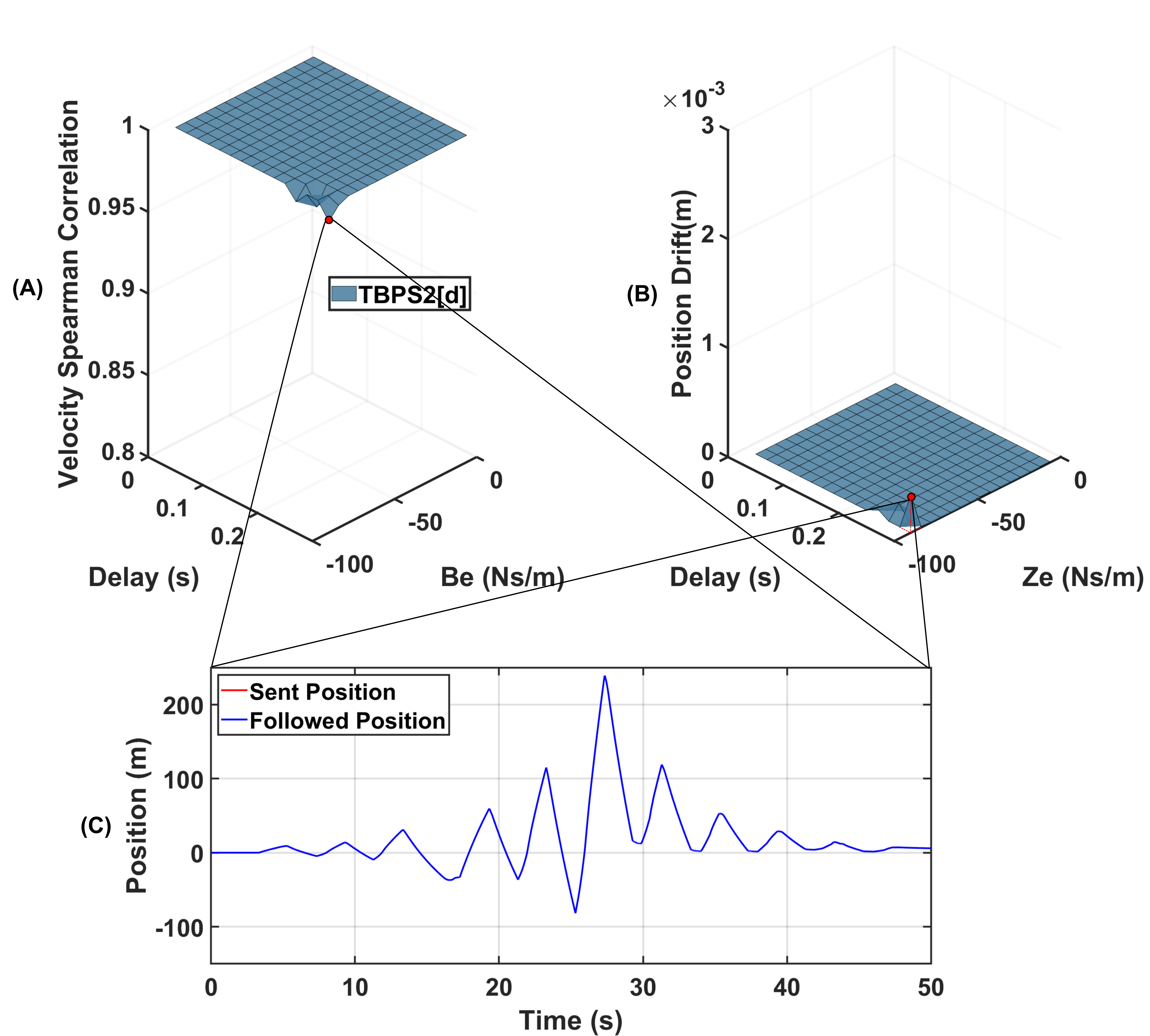}} 
\caption{(A) Spearman Correlation of the velocity tracking for TBPS\textsuperscript{2}. (B) Position drift between Leader and Follower for TBPS\textsuperscript{2}. (C) Case simulation from the assistive environment, showing the position tracking of the system under TBPS\textsuperscript{2}.}
\label{fig14}
\end{figure}

\subsubsection{Spearman Correlation for Velocity}

As can be seen in Fig. \ref{fig14}(A), the Spearman correlation between $v_2$ and $v_3$ has a consistent behavior and remains approximately 1 for all simulated values of $B_e$ and time delay. The result shows that irrespective of the Delay-$B_e$ grid setting values, the velocity on the follower side is minimally modified.


\subsubsection{Position drift during the assistive environment}

Fig. \ref{fig14}(B) showed the averaged position drift between the leader and the follower robots over the assistive Delay-$B_e$ grid simulation. The result showed that the position drift for TBPS\textsuperscript{2} is generally maintained at a significantly small value. The maximum position drift reaches around 0.03 m at the maximum delay.


To demonstrate the position tracking the behavior of the TBPS\textsuperscript{2} in assistive environments, a 50 seconds case study is given where the simulated parameter set is (delay = 300 ms, $B_e$ = -90 Ns/m), and an aperiodic sinusoidal wave signal is used as the exogenous force. The results are given in Fig. \ref{fig14}(C). As can be seen in the figure, the proposed algorithm secured a high position synchronization which can be observed by negligible position tracking error. The comparison with state-of-the-art TDPA-PS and state-of-the-art controllers cannot be conducted for non-passive environments since the conventional systems will be unstable in such scenarios. This is because of the violation of the assumption of environmental passivity. As mentioned before, this assumption is relaxed in the design of the proposed TBPS\textsuperscript{2} stabilizer.

\begin{figure}[http]
\centerline{\includegraphics[width=0.48\textwidth]{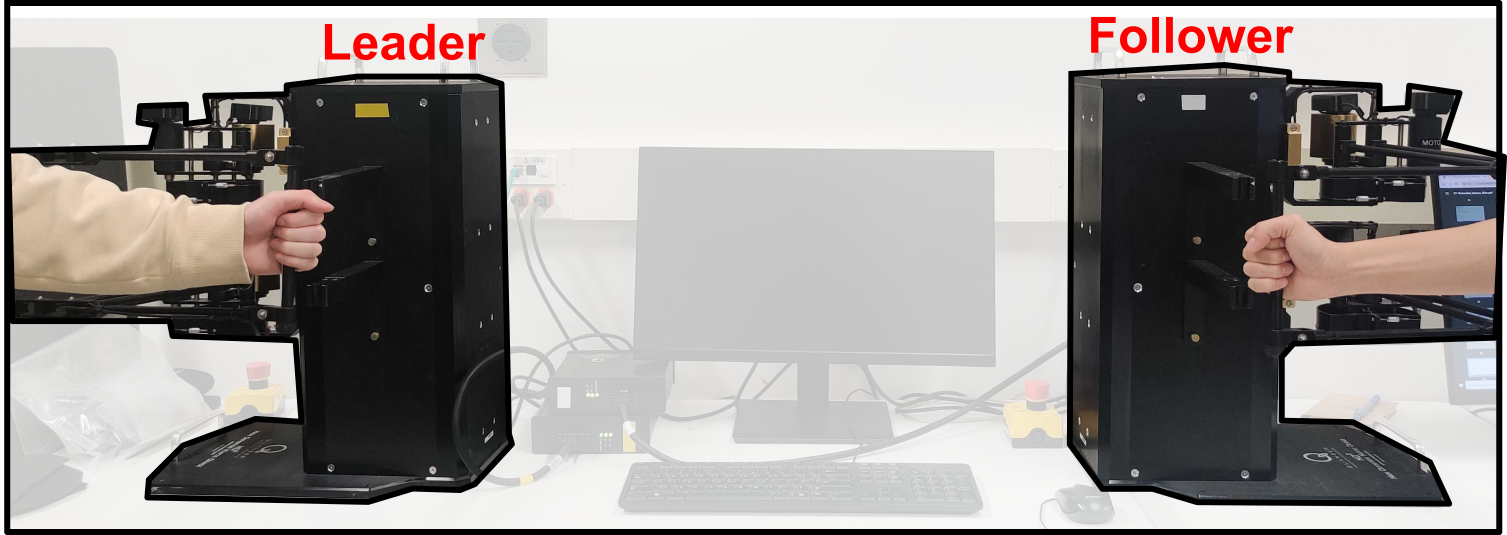}}
\caption{Experimental Set-up using two 6-DOF High-Definition Haptic Devices from Quanser Inc. One user operates the leader robot (left) while the another user operates the follower robot (right).}
\label{fig17}
\end{figure}

 \section{Experimental evaluation}
In this section, we conducted a comprehensive experiment to evaluate the practical effectiveness of the proposed stabilizer TBPS\textsuperscript{2} relative to the state-of-the-art TDPA-CF, which serves as a benchmark for TDPA designs. 
{The experimental setup incorporated a pair of 6-DOF high-resolution haptic devices manufactured by Quanser Inc (Markham, ON, Canada) and a real-time control library\cite{quanser}, QUARC, allows us to use the intuitive Simulink graphical interface to design our control system and interact with the robot in real-time.}
 The leader agent was positioned on the left, and the follower agent was positioned on the right, as depicted in Fig. \ref{fig17}. Both robotic agents demonstrated the ability to move in the XYZ Cartesian coordinate system, encompassing both translational and rotational motions having the ability to execute a diverse array of tasks. During the experiment, each robot end-effector's linear and angular velocities, force, and torque were simultaneously recorded.

\subsection{Experiment Scenario and Results}

The experiment is conducted by two operators. Operator 1 was instructed to grasp the leader robot's handle, while Operator 2 was tasked with holding the follower robot's handle. During the experiment, Operator 1 was instructed to move the robot handle sequentially along the X, Y, and Z axes, followed by rotations around the X and Y axes. Each motion lasted for 10 seconds. The follower robot's handle moved in sync with the leader robot's handle due to the teleoperation properties. While conducting this teleoperation task, operator 2 is instructed to simulate a physical obstacle, applying resistance to the follower robot's end effector in each direction's movement. Throughout the experiment, the communication delay between the two robots was modeled as a sinusoidal function, {with a constant of 100 ms and sinusoidally oscillating with an amplitude of 20 ms} ($100+20sin(t)$ ms) to assess the system's performance under varying delay conditions. The robotic system's sampling frequency was set at 1000 Hz. Furthermore, the operator's hand lower-bound EoP was estimated to be 6.8 Ns/m, determined through an offline identification procedure detailed in \cite{atashzar2017grasp}.

As can be seen in Fig. \ref{17}, the TBPS\textsuperscript{2} stabilizer exhibits a high degree of overlap between the blue and red lines, which represent the desired and actual positions, respectively. 
This phenomenon indicates that the proposed controller's position tracking performance is nearly flawless in terms of both translational and rotational movements. This is due to the novelty integration of the biomechanic energy margin into the stabilizer. Because of that, the stabilizer system possesses enhanced energy absorption capacity during the interaction, allowing it to be less conservative and achieve superior position tracking performance.

However, as can be seen in Fig.\ref{18}, for the benchmark TDPA-CF stabilizer, the red and blue lines diverge after a certain period of movement, indicating that position drift exist in the teleoperation system.
The reason for this is due to the variable time delay present in the communication channel which injects extra energy into the system and necessitates the activation of the TDPA-CF stabilizer to dissipate these excess energies to guarantee the system is passivity and, therefore stability. Although this activation can enable the telerobotic system to maintain stability during high communication delay scenarios, it comes at the cost of accumulated velocity error and, therefore, position drift within the system.

\begin{figure}[http] \label{fig18}
\centerline{\includegraphics[width=0.53\textwidth]{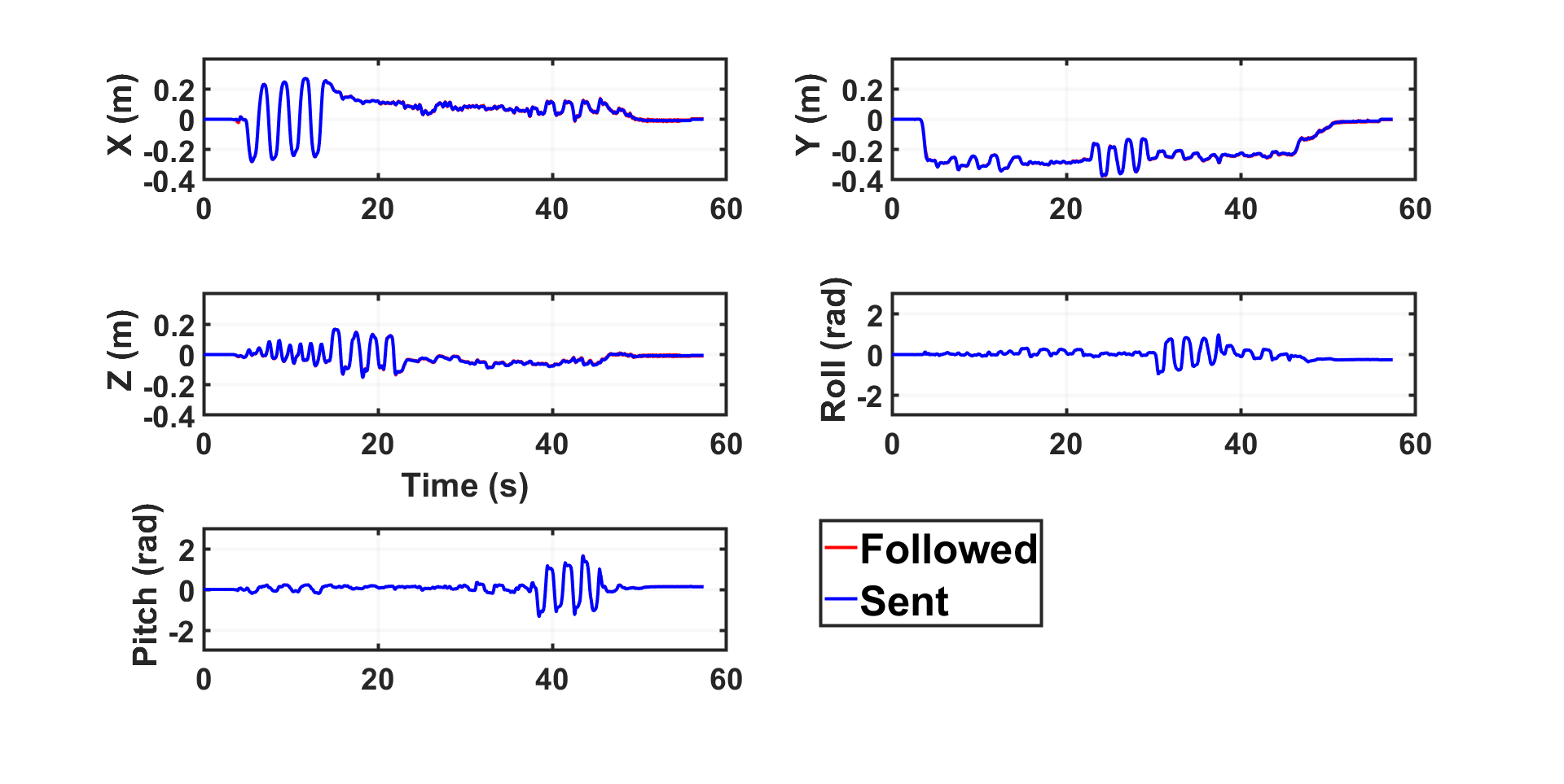}}
\vspace{-0.5cm}
\caption{Position Tracking of $TBPS^2$ in each direction}
\label{17}
\end{figure}

\begin{figure}[http]
\centerline{\includegraphics[width=0.53\textwidth]{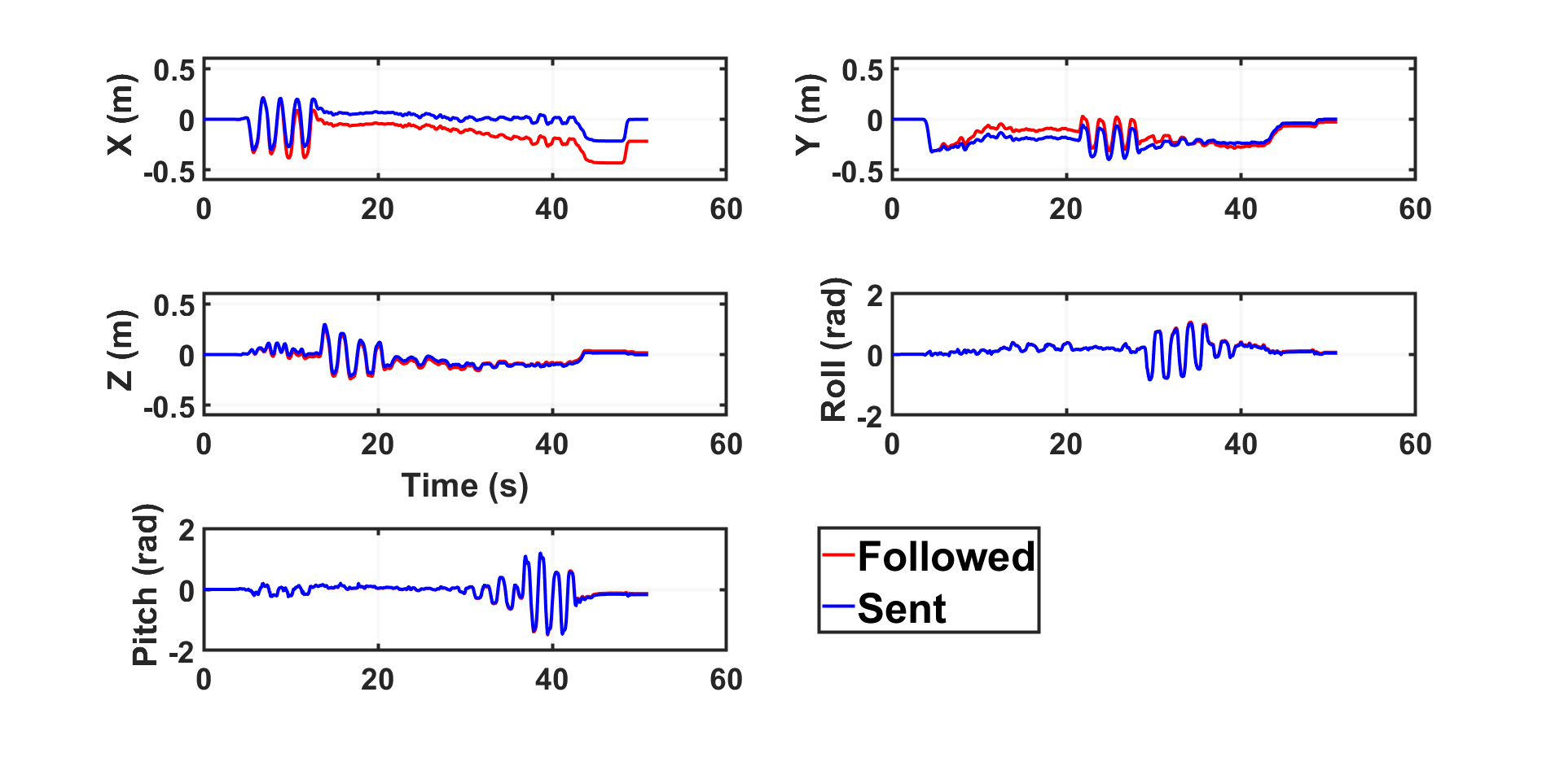}}
\vspace{-0.5cm}
\caption{Position Tracking of TDPA-CF in each direction}
\label{18}
\end{figure}

\begin{figure}[http]
\centerline{\includegraphics[width=0.53\textwidth]{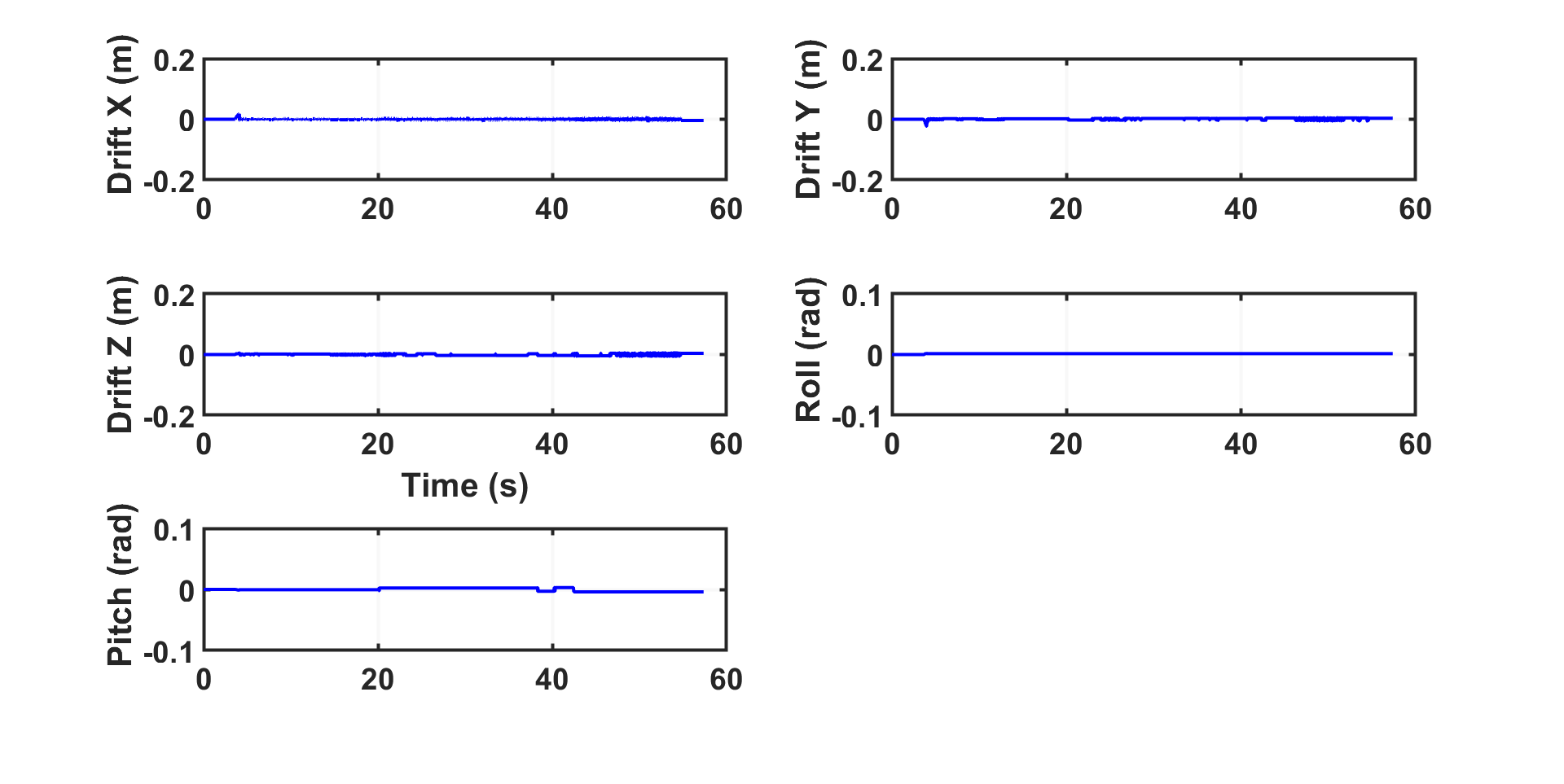}}
\vspace{-0.5cm}
\caption{Position Drift of $TBPS^2$ in each direction}
\label{fig19}
\end{figure}

\begin{figure}[http]
\centerline{\includegraphics[width=0.53\textwidth]{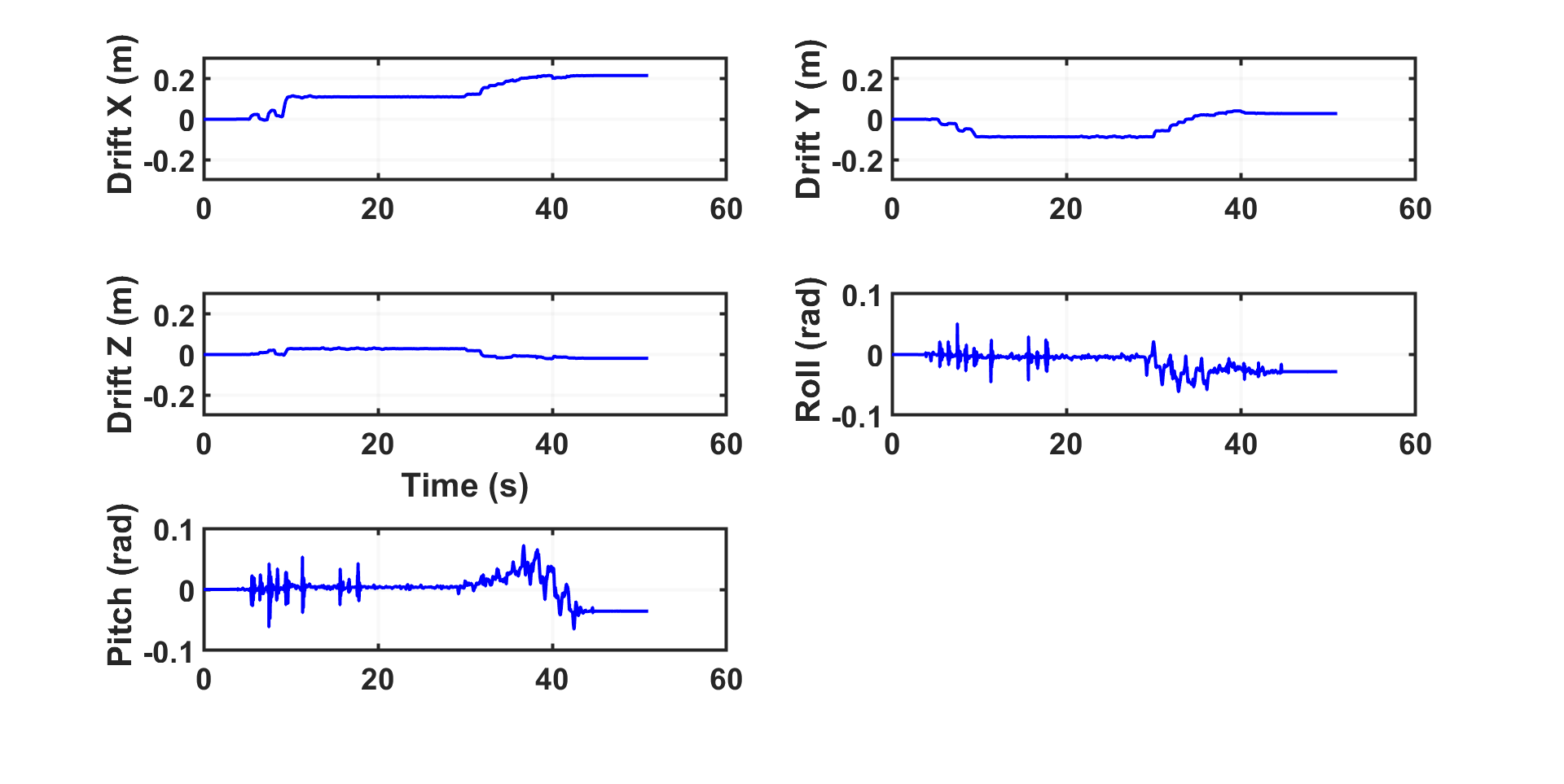}}
\vspace{-0.5cm}
\caption{{Position Drift of TDPA-CF in each direction}}
\label{fig20}
\end{figure}

In order to better demonstrate the stabilizer behavior, the position drifts of both stabilizers were measured in every movement direction, and the results are presented in Fig. \ref{fig19} and Fig. \ref{fig20}, respectively.
The $TBPS^2$ stabilizer's position drift in each direction is shown in Fig. \ref{fig19}. It is clear that there is very little position drift and that the error fluctuates between small negative and positive values.
This phenomenon is a result of the proposed stabilizer's synchronization behavior, which involves injecting or dissipating energy to accelerate or decelerate the trajectory while depending on the biomechanical energy margin to account for position drift and maintain stability. As a result, the stabilizer will activate to minimize the position drift with the given energy.
The position drift of the TDPA-CF stabilizer is presented in Fig.\ref{fig20}. As can be seen, the drift continues to increase, mainly as a result of the conservative stabilizer activation. Unlike the $TBPS^2$ stabilizer, The TDPA-CF stabilizer lacks a mechanism to account for position drift. Therefore, the accumulation of the velocity error due to the activation of the stabilizer results in a significant position drift, which may compromise the accuracy needed for the teleoperation task.

\section{Conclusion}

In this paper, a passivity-based stabilizer is proposed for networked robotic systems, named TBPS\textsuperscript{2}, as a new biomechanics-aware stabilization technique. The proposed stabilizer takes into account the energetic behavior of human biomechanics during interaction with the robotic system. Thus it integrates the biomechanical energy margin into the distributed two-port stabilization scheme. The performance of the proposed stabilizer has been mathematically proven in this paper and its practicality was investigated by a series of comprehensive simulations, followed by a systematic experiment. The results show that after the integration of the biomechanical energy margin into the proposed stabilizer architecture, the system has a more significant energy margin to maintain passivity and thus significantly improves motion tracking performance. At the same time, the proposed method mitigates the activation of the stabilizer and thus velocity deflection while guaranteeing minimal L2 stability of the system. The results also show that the TBPS\textsuperscript{2} stabilizer can relax the restrictive classical assumptions on linearity and passivity of the ports and guarantee stability without restriction on the passivity behavior of the environment and communication delay. Due to the efficient performance of the proposed stabilizer in various scenarios demonstrated in this paper, the stabilizer could be employed in various telerobotic applications, such as telerehabilitation, telesurgery, and space.

\bibliographystyle{unsrt}

\bibliography{reference}

\end{document}